\documentclass[11pt]{article}
\usepackage[utf8]{inputenc}
\usepackage{graphicx}
\usepackage{amsmath}
\usepackage{amssymb}
\usepackage{multirow}
\usepackage{amsthm}
\usepackage{wrapfig}
\usepackage{comment}
\usepackage{graphicx}
\usepackage{bm}
\usepackage[english]{babel}
\usepackage{natbib}
\usepackage{csquotes}
\usepackage[section]{placeins}
\usepackage{floatrow}
\usepackage{soul}
\usepackage{siunitx}
\usepackage[skip=2pt]{caption}
\usepackage[ruled]{algorithm2e}

\DeclareCaptionFormat{myformat}{\fontsize{6}{6}\selectfont#1#2#3}
\usepackage{xcolor}
\newcount\Comments  % 0 suppresses notes to selves in text
\definecolor{darkgreen}{rgb}{0,0.6,0}
\newcommand{\kibitz}[2]{\ifnum\Comments=1{\color{#1}{#2}}\fi}
\newcommand{\rmr}[1]{\kibitz{red}{[RESHEF: #1]}}
\newcommand{\tbs}[1]{\kibitz{blue}{[TSVIEL: #1]}}
\newcommand{\David}[1]{\kibitz{orange}{[DAVID: #1]}}

\def\calL{\mathcal{L}}
\def\calR{\mathcal{R}}

\def\XX{\boldsymbol{X}}

\newcommand{\alg}{\textsc{A}}
\newcommand{\algv}{\textsc{A}^\sigma}
\newcommand{\ebBlue}{\textsc{EbBlue}^\sigma}
\newcommand{\ebAlg}[2]{\textsc{Eb}_{\textsc{#1}}^{#2}}
\newcommand{\estm}{\phi}
\newcommand{\estmv}{\phi^\sigma}
\newcommand{\EB}{\estm_{\textsc{EB}}}
\newcommand{\EBv}{\estmv_{\textsc{EB}}}
\newcommand{\estn}{\psi}

\def\cite{\citep}

\def\newpar{\vspace{-0mm}\paragraph}

\newtheorem{theorem}{Theorem}[section]
\newtheorem{corollary}{Corollary}[theorem]
\newtheorem{lemma}[theorem]{Lemma}
\newtheorem{proposition}[theorem]{Proposition}
\newtheorem{definition}{Definition}[section]
\begin{document}
\title{Empirical Bayes approach to Truth Discovery problems}
\author{Tsviel Ben Shabat, Reshef Meir, David Azriel\\Technion---Israel Institute of Technology}
\maketitle

\begin{abstract}
When aggregating information from conflicting sources, one's goal is to find the truth. Most real-value \emph{truth discovery} (TD) algorithms try to achieve this goal by estimating the competence of each source and then aggregating the conflicting information by weighing each source's answer proportionally to her competence. However, each of those algorithms requires more than a single source for such estimation and usually does not consider different estimation methods other than a weighted mean. Therefore, in this work we formulate, prove, and empirically test the conditions for an Empirical Bayes Estimator (EBE) to dominate the weighted mean aggregation. Our main result demonstrates that EBE, under mild conditions, can be used as a second step of any TD algorithm in order to reduce the expected error.
\end{abstract}
%\David{\st{I would add that the EBE can be considered as a second step of any TD algorithm and when applied to any given TD algorithm the MSE is never worse under some conditions.}}
%\rmr{names appear as [first name] [last name]}

\section{Introduction} %\rmr{maybe start with the famous story of Galton and the ox?}
During the early 20-th century Sir Francis Galton, the 84 year old  polymath, stumbled upon a prize winning contest where approximately 800 people paid a small fee to try and guess the weight of the presented live ox after it were to be slaughtered and dressed~\cite{voxPopuli}. While no one had guessed the exact weight, Sir Francis noticed that the median guess had a negligible error. This seminal demonstration of "wisdom of the crowds" is an instructive example of \emph{truth discovery}.

In a typical instance of \emph{truth discovery} a group of workers answer questions that have correct yet unknown answers.   One  such  question  could  be,  ``What  is  the  height  of  the  building  in this image?” 
The workers who answer this question could be ordinary people, trained volunteers, a panel of experts, different computer algorithms, or a mix of all the above; all of whom we refer to as workers.
One problem is that some workers are better than others in some tasks, and some algorithms are better than others for different kinds of data. For example, if we know that worker \emph{A}, is usually better at estimating buildings' heights than worker \emph{B},
we would incorporate this fact into estimating the correct answer. For example, one can assign a different weight to different workers; in this example of the weighted mean method, it can be decided that the opinion of worker \emph{A} weighs the same as the opinion of two workers. Like so, it can also be decided that the opinion of worker \emph{B} weighs less than the opinion of a single worker, placing the weighted mean closer to the opinion of worker \emph{A}. 

%\rmr{there should be a macro that puts citations in brackets, see e.g. natbib package} 
\citet{aitkin1935least} provided a formal solution to the problem of estimating the answer of a single numerical question, answered by multiple heterogeneous workers:  if we know how competent each worker is, then is it optimal to weigh the worker proportionally to her level of competence (we will later explain these terms more formally). There are numerous works on estimating workers' competence which we discuss in Section~\ref{sec:relatedwork}.

For a single worker answering multiple questions, it may seem unlikely that there is any better solution than simply following the worker's answers. Yet,  the Empirical Bayes approach shows that a better solution indeed exists~\cite{stein1956inadmissibility,Casella85}.

In this paper we  address the following question:\\\\
    \textbf{Suppose we have \emph{multiple questions} answered by \emph{multiple heterogeneous workers}. 
    Can the Empirical Bayes approach be exploited to improve upon existing truth-discovery algorithms?}

We explore this question both when workers' competence is known (in which case our baseline is Aitkin's estimator); and when the competence is \emph{estimated}, using  an arbitrary unbiased estimator of the true answers as baseline.

In Section~\ref{sec:prelim} we formally cover the background we discussed above, that will be needed for the rest of the paper (in particular BLUE and Empirical Bayes). 

We then turn to answer the question above using a simple principle: first aggregate workers' answers, and then apply an Empirical Bayes estimator on the outcome to improve it. 

In the case of multiple workers with known competences who answer multiple questions (Section~\ref{sec:known}), we prove that our combined algorithm dominates the best linear unbiased estimator of \citet{aitkin1935least}.

In the more general setting where workers' competences are unknown (Section~\ref{sec:unknown}), we combine the Empirical Bayes estimator with an \emph{arbitrary} (unbiased) truth discovery algorithm, and characterize an exact condition under which the combined algorithm improves upon the base algorithm. These results may also be of interest outside the truth-discovery domain, as an extension of the Empirical Bayes method to situations where the variance is estimated. 

In Section~\ref{sec:emp}, we demonstrate the benefit of the Empirical Bayes approach on synthetic and real datasets, by showing how it improves upon various truth-discovery algorithms  from the literature that are used as a black-box, and discuss the practical conditions for such improvement. 

Finally, we compare our work to some related papers in the truth discovery literature and discuss implications. 

\section{Preliminaries}\label{sec:prelim}
Throughout the paper, we assume a set of $n$ workers provide answers to $m$ real-valued questions. 
The notation $\vec 1$ denotes an $m$-length vector where all entries are `1'. 
For an $m$-length vector $\vec v$, we denote its mean by $\bar v$. An $n\times m$ matrix is denoted by a bold uppercase letter (e.g. $\XX$).
%\rmr{The preliminaries section presents all the formal background that you need to later present your own results. In this case BLUE and 1-worker EB. The parts that you added move to the next section.}
\subsection{Model and Notation}
\newpar{Noise Model}  Unless mentioned otherwise, we assume workers answers follow the 
\emph{additive white Gaussian} (AWG) noise model; see \citet{diebold1998elements}. Specifically, for a worker with variance $\sigma^2$, and a question whose true answer is $\mu$, the answer is a random variable sampled from the Normal distribution $\mathcal N(\mu,\sigma^2)$. That is, workers with lower $\sigma$ are more accurate. 

%\rmr{The notation $X$ (with various subscripts) is used throughout as a scalar, row-vector, column-vector and matrix. I suggest to use different notation as in the IJCAI submission. You already do this to some extent with $\bar X$.  We can discuss this.}

%Moreover, 
%we assume that the true answers,  denoted $(\mu_j)_{j\leq m}$, are sampled i.i.d. from some distribution $\mathcal H$.
%\David{I don't think this assumption is used. In particular the expectations below are for fixed $\mu$'s. I suggest to remove this assumption. It is a motivation for the empirical Bayes estimators but the statistical analysis is for fixed $\mu$. }
 %\rmr{any assumption on the distribution of $\sigma$?}

%\David{Notice that $\sigma^2$ has two meanings in your notation. In (1) it means the competence of a single worker and in (2) it is the vector of variances. Maybe you can use boldface to vectors or $\vec{\sigma}^2$ or something else, but you cannot have the same notation for two different things}

%\David{\st{Please notice: I changed back the notation. I don't have a good solution and thinking about it again, the inconsistency in the notation seems to me minor. }} 

\newpar{Observations} The response of the $i$-th worker to the $j$-th question is denoted by $X_{ij} \sim \mathcal{N}(\mu_j, \sigma_i^2)$; it is assumed that the responses are independent. Our goal is to estimate the $m$ unknown \emph{ground truth} (GT) answers $\vec{\mu} = (\mu_1,..,\mu_m)$, We denote by  $\vec{{\sigma}}^2 = (\sigma_1^2,\dots, \sigma_n^2)$ the vector of workers' variances, where $\sigma^2_i$ is referred to as the inverse of the $i$-th workers' competences. It follows that competent workers have low variance and vice versa. %\rmr{maybe use a different term since low variance means high competence}\tbs{changed it to variance}\rmr{you can say here that `worker's' competence is just the inverse variance.}

%\rmr{maybe find a more interesting example than building heights? different sources for populations of cities? weatherpeople forecasting temperature? change in stocks' price? }\tbs{I'm keeping this comment alive in case I'll have the time to do it}
As a concrete example, one can think about the observations as a crowd-sourcing task where workers are presented with images of buildings and are told to estimate their heights. The number of workers is $n$ and the number of building images is $m$. The quantity $X_{ij}$ is the i-th worker estimate of building image $j$ and we wish to estimate the buildings' true heights  $\mu$.
The matrix of all responses is denoted by $\XX \in \mathbb{R}^{n \times m}$ (with no subscript), the answers of $n$ workers to the j-th question is denoted by $\vec{X}_j = (X_{1j},\dots,X_{nj})$
%\rmr{maybe use $\boldsymbol{X}$ to denote matrices}
, and a dataset of a single worker as $\vec X\in \mathbb{R}^m$ and her variance is denoted by $\sigma^2$.
%\David{So now $X$ has two meanings: the matrix of all responses and a single worker. Can you remove the latter? Why it is needed?}

%\David{In the following, I think that $\mu$ should be $\vec{\mu}$. Please check. I think that this replacement should be made in many places throughout the paper. Also $\sigma$ should be $\vec{\sigma}$. Please check.}\tbs{I think specifically in the dist. notations we left it without the arrow sign, I agree it could be confusing} \David{I suggest changing it. It is easy, you just need to look for `\_\{mu,sigma\}', and change all}
We denote by $\mathcal{P}_{\vec{\mu}, \vec{\sigma}}$   the distribution of $\XX$ under the parameters $\mu,\sigma$, where $\vec{\mu} \in \mathbb{R}^m, \vec{\sigma} \in \mathbb{R}_+^n$, i.e, $\boldsymbol{X} \sim \mathcal{P}_{\vec{\mu}, \vec{\sigma}}$ (and $\mathcal{P}_{\vec{\mu}, \vec{\sigma}}$ follows the AWG model unless stated otherwise) . We denote by  $E_{\vec{\mu}, \vec{\sigma}} [\cdotp]$ and  $Var_{\vec{\mu}, \vec{\sigma}} [\cdotp]$ the expected value and variance of the term in brackets, respectively, for given parameters. That is, $E_{\vec{\mu}, \vec{\sigma}}[\cdot]$ is a shorthand for $E_{\boldsymbol{X} \sim \mathcal P_{\vec{\mu},\vec\sigma}}[\cdot]$ and likewise for the variance. %\rmr{that is, $E_{\vec{\mu}, \vec{\sigma}}[f(X)]$ is a shorthand for $E_{X\sim \mathcal P_{\vec{\mu}, \vec{\sigma}}}[f(X)]$? I'm not sure it is the best notation as often the subscript of $E$ is the random variable.} \rmr{For the unknown competence part we can replace $E_{\vec{\mu}, \vec{\sigma}}$ with $E_{\mu,\calF}$ where $\calF$ is the distribution over competence levels.}\tbs{I didn't assume a distribution over the competences because we don't need it, instead we only assume that it exists and estimated}\rmr{ok I'll check but then the risk (for example) depends both on the real $\sigma$ and the estimated $\sigma$}
\newpar{Algorithms}
A \emph{truth discovery algorithm}  is a function $\alg: \mathbb{R}^{n\times m}\rightarrow \mathbb R^{m}$, mapping an observation matrix to a vector of estimated answers.\footnote{This is sometimes called an estimator but since we consider various types of estimators in this work, we use the term truth discovery algorithm to avoid confusion.}  
An algorithm may also take additional information as input. In particular, a \emph{variance-based algorithm} (denoted by $\algv$) is assumed to have access to the true variance of each worker. 

\newpar{Evaluation} Given a \emph{truth discovery algorithm}, we are interested in how far $\alg(\XX)$ is from the true answers $\vec \mu \in \mathbb{R}^m$, in expectation.  

Formally, we denote by
 $\mathcal{L}(\hat \mu, \vec{\mu})$  the \emph{loss} of estimation $\hat \mu \in \mathbb{R}^m$. Throughout this work, the loss function $\mathcal{L}$ is the square euclidean norm, i.e, $\mathcal{L}(\hat \mu, \vec{\mu}) := \| \hat\mu - \vec{\mu} \|_2^2$. We then measure the loss of $\alg$ on a particular input as $\calL(\alg(\XX),\vec{\mu})$.
 
  Because the observations $\boldsymbol{X}$ are random, we will use the expected loss. Formally, the \emph{risk} of Algorithm $\alg$ (given parameters $\vec{\mu},\vec{\sigma}$) is %\rmr{$\vec \sigma$ and $\vec\mu$?}
  $$\mathcal R_{\vec{\mu}, \vec{\sigma}}(\alg) := E_{\vec{\mu}, \vec{\sigma}}[\calL(\alg(\XX),\vec{\mu})].$$

  Our goal then is to find some algorithm $\alg$ with low risk, i.e, to minimize $\mathcal{R}_{\vec{\mu}, \vec{\sigma}}(\alg)$ for every $\vec \mu$ and $\vec\sigma$.

\subsection{The Best Linear Unbiased Estimator (BLUE)}
% We shall use the following definitions. 
% \begin{definition}
Recall that an estimator of a parameter is:
\begin{itemize}
    \item \emph{unbiased} if its expected value equals the estimated term;
    \item \emph{linear} if it is a linear function of the observations.
\end{itemize}
% An unbiased estimator $\delta$ of a parameter $\mu$ is an estimator whose expected value is equal to the parameter it is estimating. i.e., $E_{\vec{\mu}, \vec{\sigma}}(\delta(\boldsymbol{X}, \sigma)) = \vec{\mu}$ for all $\mu$. 
% %\David{Do you mean $E_{\vec{\mu},\vec{\sigma}}(\delta({\bf X},\vec{\sigma})) = \vec{\mu}$ for all $\vec{\mu}$ and $\vec{\sigma}$ ?? If so, pay attention to the notation!}
%\end{definition}
 \begin{table}
 %\begin{minipage}{.5\linewidth}
 \begin{scriptsize}
 \begin{center}
  $ \begin{array}{|c|cccc|c|}
    \hline  
  X_{ij}&  1 & 2 & 3& 4 & \sigma_i^2 \\ \hline
1	&		20	&	2  & 3 & 4 & 93.5	\\
2	&		10	&	11 & 18 & 14	& 11\\
3	&		8	&	11 & 23 & 19 & 34.5\\
4	&		6	&	13 & 7 & 3  & 56.5 \\
											\hline\hline 
GT & 10 & 9 & 12 & 16 & \mathcal{L}  \\
\hline
AVG & 11 & 9.25 & 12.75 & 10 & 9.41 \\		
\algv_B & 9.85	&	10.6	&	16.6	&	12.95 & 8.22 \\
\ebBlue & 10.25 & 10.82 & 15.5 & 12.65 & 6.68 \\
 \hline
%OA(S)&&8.31&-0.40&3.80&2.13&0.01\\
    \end{array}$
    \end{center}
    \end{scriptsize}\vspace{-2mm}
    \caption{An example of a data set, $X_{ij}$ is the i-th worker response for the j-th question. $\sigma^2_i$ is the calculated variance of the i-th worker relative to the ground truth (GT), $\mathcal{L}$ is the loss of each estimator }\label{table:example_ds}
    \end{table}
% \begin{definition}
% A linear estimator $\delta$ is a linear function of the observable random variables.
% \end{definition}
% We denote  by $\Delta_{LUE}$ the set of all linear unbiased estimators. A simple example of an estimator in $\Delta_{LUE}$ is the unweighted average  $(\frac{1}{n}\sum_{i=1}^n X_{i1}, \ldots, \frac{1}{n}\sum_{i=1}^n X_{im})$. 
% %\David{I think you mean something else: $(\frac{1}{n}\sum_{i=1}^n X_{i1}, \ldots, \frac{1}{n}\sum_{i=1}^n X_{im})$. Please check!}

We denote by $\Delta_{LUE}$ the set of all linear unbiased estimators $\algv$, i.e. all  linear unbiased truth discovery algorithms that also have access to workers' variance. 

%Denote the estimator $\delta_B: \mathbb R^n \times \mathbb R_+^n \rightarrow \mathbb R$, where  %\rmr{subscript or superscript?}
Consider the following estimator/algorithm:
$$ \algv_B(\vec X_j) :=  \big(\sum_{i=1}^n\frac{1}{\sigma_i^2}\big)^{-1}\sum_{i=1}^{n}\frac{X_{ij}}{\sigma_i^2}.$$
and $\algv_B(\XX):= (\algv_B(\vec{X}_j))_{j\leq m}$. It can be easily shown that $\algv_B$ is an unbiased estimator i.e $E_{\vec{\mu}, \vec{\sigma}}[\algv_B(\XX)] = \vec{\mu}$.

\begin{theorem}[\cite{aitkin1935least}]
\label{thm:Aitkin}
Under the AWG model, 
 $\calR_{\vec{\mu}, \vec{\sigma}}(\algv) \geq \calR_{\vec{\mu}, \vec{\sigma}}(\algv_B)$, for all $ {\algv} \in\Delta_{LUE},  \vec\mu \in \mathbb{R}^m, \vec\sigma \in \mathbb{R}^n_+$. %\rmr{need to say this holds for all $\mu,\sigma$. Also I don't understand why $Var({\delta}(X)) - Var(\delta_B(X))$ is a matrix?}
%\rmr{better to use the loss in the risk in the theorem. I guess that since $\delta,\delta_B$ are unbiased then $Var_{\vec{\mu}, \vec{\sigma}}({\delta}(X,\sigma)) = \calR_{\vec{\mu}, \vec{\sigma}}(\delta)$, right?}
\end{theorem}
In words, the theorem of \citet{aitkin1935least} shows that the inverse variance weighing of the observations is the best linear unbiased estimator (BLUE) for $\vec{\mu}$ under the square loss function. In particular, the BLUE uses the input on each question separately, and thus the risk is independent of the number of questions $m$. As per our buildings' heights example, if we know how competent each worker is (at estimating buildings' heights from images), the best unbiased linear way of estimating the real height is a weighted average of the workers' answers, where the weight of worker~$i$ is $1/\sigma_i^2$.

Table \ref{table:example_ds} is an example for a dataset, where workers' variances are known and the loss for this particular instance  is compared between different estimators.

\subsection{The Empirical Bayes Estimator (EBE)} %\rmr{when there is only one scalar $\sigma$ I think the $\bar\sigma$ is redundant}

%\rmr{We can avoid confusion if we use a different letter for estimators that take a single $m$-size vector. E.g $\phi_{EBE}$ instead of $\delta_{EBE}$}\tbs{Well EBE and $\delta_I$ always takes single m-size vector and the rest don't, I remember that one of our past reviews for proxy complained about too much notation, What about $\boldsymbol{X}$ and $\vec{X}$ doesn't it solve the problem? I'm so busy with changing notations all over the text and I didn't really get to Section 5 yet}
%\rmr{I don't understand what is $\bar X$, what is $\vec X$ and what is $\bar X \mathbf{1}$.}
In a seminal paper, \citet{stein1956inadmissibility} introduced an estimator that dominates BLUE in a simple estimation problem of a normal distribution. In the current setting, in the case of a single worker estimating multiple questions, Stein's result implies that estimating the true answers using the input of the worker directly, is dominated by another estimator, which is rather surprising. %. Stein's result is surprising, since using $\delta_I$ is intuitive and commonly used. 
Consequently, different variations of Stein's estimator and its derivation from  empirical Bayesian statistics perspective are introduced by \citet{10.2307/2284155}. Here we focus on one such version.\\
To define the estimator, we consider a setting of a single worker with responses $\vec{X}=(X_1,\ldots,X_m)$, following a single-worker AWG model. That is, $X_1,\dots,X_m$ are independent with $X_j \sim \mathcal{N}(\mu_j, {\sigma}^2)$, $j=1,\ldots,m$. 

A \emph{modifying estimator} is a function $\estm:\mathbb{R}^m \times\mathbb{R}_+\rightarrow \mathbb R^m$, which can `modify' a vector of responses. The estimator can also accept an additional parameter, which we can think of as the (true or estimated) variance. As with truth discovery algorithms, we denote by $\estmv:\mathbb{R}^m \rightarrow \mathbb R^m$ modifying estimators that have access to the true variance $\sigma^2$. 

A trivial example is the identity estimator $\estm_{I}(\vec X):=\vec X$. 
%\David{I suggest adding something like: To define the estimator, we consider a setting of a single worker with responses $\vec{X}=(X_1,\ldots,X_m)$. It is assumed that $X_1,\dots,X_m$ are independent with $X_j \sim \mathcal{N}(\mu_j, \bar{\sigma}^2)$, $j=1,\ldots,m$.}

%\rmr{the definition is independent of the assumptions, which are irrelevant here. Also this is an estimator so why not use the estimator notation? For a single worker with observations $\vec X=(X_1,\ldots,X_m)$ and variance $\sigma^2$ (a scalar?) we define:
%$$\delta_{EBE}(\vec X,\sigma) :=  \bar{X}{\boldsymbol{1}} + %\Big[1 - \frac{(m-3)\bar{\sigma}^2}{ \|\vec{X} -\bar{X}{\boldsymbol{1}}\|^2}\Big](\vec{X}-\bar{X}{\boldsymbol{1}}).$$}
\begin{definition}[Empirical Bayes Estimator]\label{EBE}
%\rmr{I think the words `their variance' implies that $\sigma$ is the sample variance of $\vec X$. since you already defined it correctly above, better to just remove it from here}
%Given observations, $\vec{X}=(X_1,\ldots,X_m)$ and their variance, denoted by $\bar{\sigma}^2$, the EBE for $\vec{\mu}$ is
\begin{equation}\label{eq:EBE}
\EB(\vec{X}, {\sigma}) := \bar{X}{\vec{1}} + \Big[1 - \frac{(m-3){\sigma}^2}{ \|\vec{X} -\bar{X}{\vec{1}}\|^2}\Big](\vec{X}-\bar{X}{\vec{1}}).
\end{equation}
%\rmr{moved the general notation comments to the beginning}
%where $\Bar{X} = \frac{1}{m}\sum_{j=1}^m X_j$ and $\boldsymbol{1}$ is a vector of ones of size $m$.
\end{definition}
% \David{The assumptions about normality are not part of the definition. I suggest something like:
% Given observations, $\vec{X}=(X_1,\ldots,X_m)$ and their variance, denoted by $\bar{\sigma}^2$, the EBE for $\vec{\mu}$ is
% \[
% \delta_{EBE}(\vec{X}, \bar{\sigma}) = \bar{X}{\boldsymbol{1}} + \Big[1 - \frac{(m-3)\bar{\sigma}^2}{ \|\vec{X} -\bar{X}{\boldsymbol{1}}\|^2}\Big](\vec{X}-\bar{X}{\boldsymbol{1}}),
% \]
% where $\Bar{X} = \frac{1}{m}\sum_{j=1}^m X_j$ and
% $\boldsymbol{1}$ is a vector of ones of size $m$.
% }
Arranged differently, $\EB(\vec{X}, \bar{\sigma}) = \bar{X}{\vec{1}} \frac{(m-3){\sigma}^2}{ \|\vec{X} -\bar{X}{\vec{1}}\|^2} + \vec{X}\Big[\vec{1} - \frac{(m-3){\sigma}^2}{ \|\vec{X} -\bar{X}{\vec {1}}\|^2}\Big]$ is a weighted average of each component of $\vec{X}$ and its mean $\bar X$. It is also instructive to notice that when $m=3$ then $\EB(\vec{X}, \sigma) = \vec{X}$, i.e, $\EB$ becomes the identity estimator $\estm_I$, and when $m$ goes to infinity we get that $\lim_{m \to \infty} \frac{(m-3){\sigma}^2}{ \|\vec{X} -\bar{X}{\bf1}\|^2} = \frac{{\sigma}^2}{{\sigma}^2 + C}$ where $C\in \mathbb{R}_+$ is a constant related to the variance of the ground truth. We provide an explicit expression for C, and the derivation of the EBE in Appendix~\ref{proof:EBEderive}. %Before showing that BLUE is dominated by $\hat{\mu}_{EBE}$, we first recall the definition of admissibility.

\begin{theorem}[\cite{pointestimation}]\label{cassellaBound}
In the AWG model with a single worker and $m>3$ questions, 
\begin{equation}\label{eq:inadmissble}
    \mathcal{R}_{\vec{\mu},{\sigma}}(\EBv) < \mathcal{R}_{\vec{\mu},\sigma}(\estm_I) \ \text{ for all }\vec{\mu} \in \mathbb{R}^m, {\sigma} \in \mathbb{R}^+.
\end{equation}    
\end{theorem}
In words, the empirical Bayes estimator for $\vec{\mu}$, which is not linear and not unbiased (but has access to the true variance $\sigma$), has strictly lower risk than $\estm_I$. Note that for a single worker,  $\estm_I$ coincides with the BLUE $\algv_B$.

For completeness, we have provided a proof for Theorem~\ref{cassellaBound} (Appendix~\ref{proof:casella}) that follows Problem 6.1 in  \cite[Chapter~5]{pointestimation}.

We also consider Stein's estimator, which is defined now.
\begin{definition}[\cite{stein1956inadmissibility}]\label{stein_estimator}
%The Stein estimator is
\begin{equation*}
\estm_{Stein}(\vec{X}, {\sigma}) :=  \big[{1} - \frac{(m-2){\sigma}^2}{\| \vec{X} \| ^2} \big]\vec{X}
%\David{\text{I think you should denote this estimator by }\hat{\mu}_{SE}}
\end{equation*}
\end{definition}
Stein estimator can be thought of as an empirical Bayes estimator with a normal prior and where the prior's mean is known to be 0.

%\David{\st{I think it would be better to say it more simply as follows}}

% \begin{definition}[Moment Estimator]
%  A Moment Estimated Bayesian Estimator (MEBE) for $\mu$ is:
% \begin{align*}%\label{eq_system}
% \delta_{MEBE}(\vec{X}, \sigma) &= \bar{X}\bf1 + \big[1 - \frac{\bar{\sigma}^2}{\hat{\sigma}_0^2 + \bar{\sigma}^2}\big](\vec{X}-\bar{X}\bf1)
% , \hat{\sigma}_0^2=max\{\frac{1}{m}\|\vec{X} -\bar{X}{\bf1}\|^2 - \bar{\sigma}^2, \ 0\} %\David{\text{I think you should denote this estimator by }\hat{\mu}_{MEBE}}
% \end{align*}
% \end{definition}
% The MEBE is similar to the EBE estimator, which was defined in \eqref{eq:EBE} and it is motivated by the same empirical Bayes framework,  but the estimator for the variance of the prior distribution is different.

%\David{What is the advantage of the MEBE? If it always preform worse than other methods, I suggest to remove it}\tbs{For completeness? I guess it can be removed but I'd rather find a plausible reason to keep it}

%\David{\st{Can you write the MEBE in a similar way as (1) so they will be easier to compare?}}

\section{Known Competence}\label{sec:known}
We now return to the general multi-worker, multi-question setting. 

We begin by assuming that we know the competences of the workers  $\sigma_i$ for $i = 1,\ldots,n$, an assumption that will be relaxed later. %Since $\vec{\sigma}$ is fixed, we will omit the argument $\vec{\sigma}$ and write only $\delta(\boldsymbol{X})$. %In this section we first present the best linear unbiased estimator (BLUE) for estimating $\mu$, and we suggest a sufficient statistic point of view for it. We then introduce the Empirical Bayes estimator and show that it dominates the BLUE, implying that the latter is an inadmissible estimator for $\mu$.

%\rmr{here you should put whatever results you have for the known-competence model}

We will use the following result which stems from Neyman-Fisher factorization theorem.
\begin{proposition}\label{prop:suff_stat}
$\algv_B(\boldsymbol{X})$ is  a sufficient statistic for $\vec{\mu} = (\mu_1,..,\mu_m)$ under $\mathcal{P}_{\vec{\mu}, \vec{\sigma}}$.
\end{proposition} We provide a proof in appendix \ref{suff_statistic}\\
%\rmr{now if you want to switch to a simpler notation you can define e.g. $X^*_j:=\delta_B(X_j,\sigma)$ (or $\mu^*_j$).}
\par It follows that there is no loss of relevant information when considering only $\algv_B(\XX)$ instead of the observation matrix. That is, we can replace our observations $\XX \in \mathbb{R}^{n \times m}$ with a single `aggregated worker'' who answered $m$ questions%\David{(Notice the change of notation)}
, denoted by $\vec{X}^B=(X^B_1,...,X^B_m) = \algv_B(\XX)$. The variance of this single worker (denoted by $\tilde\sigma^2$) is the harmonic mean of all workers' variances, divided by $n$. %The results of this section are summarized in the following corollary.
% \begin{corollary}\label{agg_to_single}\rmr{This corollary seems redundant - it just repeats the proposition above}
% Under the model the model $X_{ij} \sim \mathcal{N}(\mu_j, \sigma_i^2) \ i = 1,..,n \ j = 1,..,m $ independently, when the workers' competence is known, there is no loss of information in using only the vector $\delta(\boldsymbol{X})$, which is the BLUE, instead of the matrix $\boldsymbol{X}$.
% \end{corollary}

%Our suggested  estimator is described now.
\begin{algorithm}[t]
\caption{$\ebBlue$ for Known Competence}\label{alg:one}
\KwIn{Dataset $\boldsymbol{X} \in \mathbb{R}^{n\times m}$, variances $\vec{\sigma} \in \mathbb{R}_+^n$}
%\KwOut{$\vec{T} \in \mathbb{R}^m$} 
 $\vec{X}^B \gets \algv_B(\XX)\ $;\\
 $\tilde{\sigma}^2 \gets \big(\sum_{i=1}^n\frac{1}{\sigma_i^2}\big)^{-1} $;\\
 \Return $\EB(\vec{X}^B, \tilde{\sigma})\ $;
\end{algorithm}

Our first algorithm $\ebBlue$ (see Alg.~\ref{alg:one}) simply uses  Aitkin's BLUE to aggregate the labels independently on each question, then applies the Empirical Bayes estimator on the outcome. Since $\tilde\sigma^2$ is the true variance of the aggregated worker, $\EB(\vec{X}^B, \tilde{\sigma})=\EBv(\vec{X}^B)$, so intuitively we are back to applying Empirical Bayes in a single-worker scenario. Indeed,  from Proposition~\ref{prop:suff_stat} and Theorem~\ref{cassellaBound} we get our main result for this section:

\begin{corollary}\label{corollary:inadm_blue}
 In the AWG model, for $m>3$ and any $n$,
 \begin{equation}\label{eq:alg1}
    \mathcal{R}_{\vec{\mu}, \vec{\sigma}}(\ebBlue) < \mathcal{R}_{\vec{\mu}, \vec{\sigma}}(\algv_B) \text{ for all } \vec{\mu} \in \mathbb{R}^m, \vec{\sigma} \in \mathbb{R}_+^n 
\end{equation}   
 %\rmr{can you write more explicitly that $R(Alg1) < R(\delta_B)$? Also, recall that the theorem/corollary does not need to repeat the assumptions if they were specified before.}
\end{corollary}
\begin{figure}
    \centering
    \includegraphics[height=4cm,width=7cm]{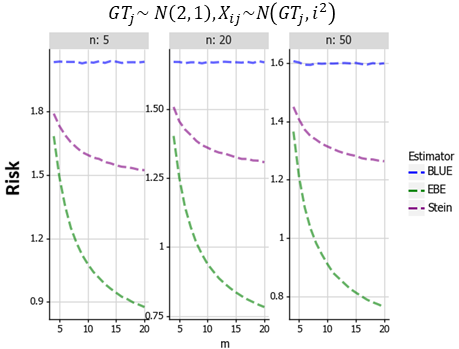}
    \captionsetup{justification=justified}
    \caption{Each data point is a 100,000 samples average, each sample includes new GT and new workers.}
    \label{figure:known_competence}
\end{figure}
In words, Corollary~\ref{corollary:inadm_blue} says that if we know the workers' competences then Alg.~\ref{alg:one} strictly beats the unmodifed BLUE, and by Theorem~\ref{thm:Aitkin} strictly beats any unbiased linear estimator. In statistical terms, BLUE (and any other unbiased linear estimator) is an inadmissible estimator.
%the best linear weighted average (BLUE) will \emph{always} have a higher risk than EBE.\\
\par Figure \ref{figure:known_competence} demonstrates inequality \eqref{eq:alg1}. We generated ground truth $\mu_j\sim N(2, 1)$, $j=1,\ldots,m$ independently, and noisy observations $X_{ij} \sim  N(\mu_j, i^2)$, %\David{How $\sigma_i$ was generated?}
 then we aggregated the answers into a single worker using $\algv_B$,  applied  different modifying estimators on the outcome, and calculated their (empirical) risk over 4 samples. It can be seen that the advantage of EBE and Stein is more significant when fewer workers answer more questions.
% \par \emph{In this section we showed that when the workers' competences are known, the best weighted average (BLUE) is a sufficient statistic for estimating the ground truth and if we wish to minimize the MSE then BLUE is also an inadmissible estimator - EBE is a better choice.}
\section{Estimated Competence}\label{sec:unknown}
Often in real world problems we have no access to the true variances of our workers, $\vec{\sigma} \in \mathbb{R}_+^n$. However we can estimate it from the observations. Many \emph{truth discovery} algorithms are doing exactly that---either in a supervised way (if we have access to the true answers of some questions) or unsupervised (by comparing workers to one another).

%Since applying the EBE estimate (Alg.~\ref{alg:one}) requires the true variance of the aggregated worker, a natural question is how is the risk affected when we only have an \emph{estimate} of the variance?
%That is, when Alg.~\ref{alg:one} is applied using some estimate $\hat\sigma$ rather than the true $\bar\sigma$.
We therefore abstract away from our Alg.~\ref{alg:one}, by assuming we only have access to an estimate of the aggregated worker's variance. 
Crucially, our analysis is oblivious as to \emph{how} the variance is estimated, or to how observations were aggregated, and may therefore apply for any truth-discovery algorithm.

Similarly to the previous section, we consider the aggregated answer vector. However rather than $\algv_B$ (which requires the actual workers' variances), we now assume an arbitrary truth discovery algorithm $\alg:\mathbb R^{n\times m}\rightarrow \mathbb R^m$ is used, together with some estimator of the variance 
    $\estn:\mathbb R^{n\times m}\rightarrow \mathbb R_+$.
    
Then, our general $\ebAlg{\alg}{\estn}$ algorithm (see Alg.~\ref{alg:two}) simply applies Eq.~\eqref{eq:EBE} to modify the output of algorithm $\alg$, using the estimated variance $\hat\sigma^2=\estn(\XX)$. %\rmr{did I get this right?}
\begin{algorithm}
\caption{$\ebAlg{\alg}{\estn}$ for estimated Competence}\label{alg:two}
\KwIn{Dataset $\XX\in \mathbb{R}^{n\times m}$}
%\KwOut{$T \in \mathbb{R}^m$} 
 $\vec X^A \leftarrow \alg(\XX)$;\\
 $\hat \sigma^2 \leftarrow \estn(\XX)$;\\
 \Return $\EB(\vec X^A,\hat\sigma)$;
\end{algorithm}

Our analysis is divided into two parts: we first
analyze the risk of Alg. \ref{alg:two} under the minimal assumption that each $X^A_j$  is an unbiased estimator of $\mu_j$ in Section \ref{sec:general}. Then, Section~\ref{sec:normal} considers the case where the answers of the aggregated worker are assumed to be normally distributed around the true answers.

\subsection{General Model}\label{sec:general}
As previously mentioned, \emph{truth discovery} algorithms typically estimate the truth by estimating workers' competence and then aggregate answers, weighing them accordingly.

In general we may not know the distribution of the aggregated answer, either since the initial observations depart from the AWG model, or because the algorithm $\alg$ is complicated or unknown. We thus relax any assumption on the input in this section, except that $\vec X^A$ is \emph{unbiased}. Thus $\vec\mu = E[\vec X^A]$.  We also denote the true (unknown) variance by $\sigma^2:=Var[\vec X^A_j]$ (identical for all $j$).
%The aggregated worker - the estimated truth (to which we apply $\delta_{EBE}$) is of unknown distribution, and therefore, {in this section we relax the previously assumed normal model}, instead, we require unbiasedness.
%\par Denote the dataset $\boldsymbol{X}\in \mathbb{R}^{n \times m}$ which is the answers of $n$ workers to $m$ questions where the actual unknown truth is denoted by $\vec{\mu}\in \mathbb{R}$, we further denote $\delta^A(\boldsymbol{X})=\vec{X}_A=(X_1^A,..,X_m^A)$ to be the truth estimation of some algorithm, We assume that $E(\vec{X}_A)=\vec{\mu}$ and $Var(X_i^A)=\bar{\sigma}^2$ i.e., is constant for all $i=1,\ldots,m$. We denote $\hat{\sigma}^2$ as general estimator of the variance $\bar{\sigma}^2$.
We next  present a sufficient condition for our Alg.~\ref{alg:two}  to have smaller risk than its baseline algorithm $\alg$ for all $\mu \in \mathbb{R}^m$.

% %\David{Alternative version:}\tbs{I replaced my version with yours}\David{Do you assume that $X_1,\ldots,X_m$ are independent?}\tbs{if you mean the output of an algorithm then no}
% \begin{definition}\label{general_EBE}
% For some $m > 3 $ a Generalized EBE (GEBE) is :
% \begin{equation}
% \phi_{GEBE}(\vec{X}_A, \hat{\sigma}^2) = \bar{X}_A{\vec {1}} + \Big[{1} - \frac{(m-3)\hat{\sigma}^2}{ \|\vec{X}_A -\bar{X}_A\vec{1}\|^2}\Big](\vec{X}_A-\bar{X}_A\vec{1})
% \end{equation}
% \end{definition}
%\David{\st{I think you should use a different notation here as this is a different estimator, maybe GEBE, and $\hat{\mu}_{GEBE}$?}} 
\begin{theorem}\label{general_model}
For any unbiased algorithm $\alg$, and $m>3$,\footnote{Since in this subsection we do not assume that the distribution of $\XX$ follows the AWG model, we do not need a parameter for the individual competence. Other than that, all definitions remain the same.}
$$\calR_{\vec\mu}(\ebAlg{\alg}{\estn}) <\calR_{\vec\mu}(\alg) \text{ for all }\vec{\mu} \in \mathbb{R}^m $$
if and only if
\begin{multline} \label{eq:cond} 
  2(m-3)\Sigma_{j=1}^m Cov\Big(X_j^A, \frac{\estn(\XX)(X_j^A-\bar{X}^A)}{\|\vec{X}^A -\bar{X}^A{\vec{1}}\|^2}\Big) 
  \\
  -(m-3)^2E_{\vec\mu}\Big(\frac{(\estn(\XX))^2}{\|\vec{X}^A -\bar{X}^A{\vec{1}}\|^2}\Big) >  0.
\end{multline}
\end{theorem}
%\David{Did you prove iff?}\tbs{It's equal to the actual risk difference}
%\David{\st{What is the result of the theorem? There is only an ``if'' statement. Also, do you still assume the model from section 3? Should $E$ be $E_{\vec{\mu}, \vec{\sigma}}$?}}
The condition in Theorem~\ref{general_model} may seem somewhat obtuse, yet we argue it may still be useful:
\begin{itemize}
\item It is easy to see that by choosing $\estn(\vec X)$ that is sufficiently close to 0, the condition holds;
    \item The condition is purely a function of the observations $\XX$, therefore it can be verified empirically, given enough samples;
    \item It leads to an improvement of the algorithm, as we explain below;
    \item Under additional assumptions on the distribution, the condition is substantially simplified and provides important intuition (Section~\ref{sec:normal}).
\end{itemize}
%Notice that  suggest a  theoretical way to estimate whether applying empirical Bayes will improve the outcome of the baseline algorithm $\alg$, since left hand-side of \eqref{eq:cond} is based entirely on the dataset. 
%\David{What do you mean? You need to explain. Also, you need to explain how do you plan to do bootstrap}\tbs{Changed to theoretically}

The main assumption of Theorem~\ref{general_model} is that the output of the truth discovery algorithm used as baseline is unbiased, i.e., $E(\vec{X}^A)=\vec{\mu}$. This assumption may not hold, when workers provide biased estimates. For example, in the buildings heights setting, if workers systematically overestimate the buildings' heights, then the assumption is violated (see proof in appendix \ref{proof:general_model}).
 %\David{\st{I suggest to elaborate here. Explain explicitly what is this practical method}}  \David{\st{I don't understand what is assumed in this section. I think you should explain here more. This is I think your main result. You should say what are the implications and why they are important}} 
 
 \paragraph{Generalizing further?}
While the modifying estimator $\EB$ we apply is optimal for a \emph{single worker}, it turns out that we can do better in the multi-worker case. 

We define the \emph{generalized EB estimator} $\EB^\alpha$ by replacing the $(m-3)$ term in Def.~\ref{EBE} with $\alpha\in \mathbb R_+$; and denote by $\ebAlg{\alg}{\estn,\alpha}$ the corresponding generalized version of Alg.~\ref{alg:two}.
\begin{proposition}\label{prop:alpha_star}
$\calR_{\vec\mu}(\ebAlg{\alg}{\estn,\alpha})$ is minimized by setting
 $$\alpha^* := \frac{\Sigma_{j=1}^m Cov\Big(X_j^A, \frac{\estn(\XX)(X_j^A-\bar{X}^A)}{\|\vec{X}^A -\bar{X}_A{\vec{1}}\|^2}\Big)}{E\Big(\frac{\estn(\XX)^2}{\|\vec{X}^A -\bar{X}^A{\vec{1}}\|^2}\Big)}.$$ 
\end{proposition}
%  \begin{table}\vspace{-5mm}
%  %\begin{minipage}{.5\linewidth}
%  \begin{scriptsize}
%  \begin{center}
%   $ \begin{array}{|cccc|ccc|}
%     \hline  
% GT_i&  X_{ij} & n & m & \Delta_* & \hat{\Delta}_* & \Delta\\ \hline
% N(0,i^2)	& N(GT_i, j^2) &10&10&	.032647 & .03237 & .031867\\
% N(i,1)	 & N(GT_i, j^2)    &10&10& 	.08159  & .07901 & .0800\\
% N(i,1)	 & N(GT_i, j^2)    &100&10&	.39129  & .39405 & .38030\\
% N(i,1)	 & N(GT_i, j^2)    &10&100&	.00109  & .00117 & .00074\\
% exp(i)	 & N(GT_i, j^2)    &10&10&	.044174 & .04338 & .04326 \\
% exp(i)	 & GT_i + exp(j)   &10&10&	.05094  & .05133 & .04809 \\
% exp(i)	 & GT_i + exp(j)   &100&10&	6.2414  & 6.2196 & 5.7418 \\
% uni(0,i) & GT_i + exp(j)   &10&10&	.36857 & .36381 & .34491\\ \hline
% %OA(S)&&8.31&-0.40&3.80&2.13&0.01\\
%     \end{array}$
%     \end{center}
%     \end{scriptsize}\vspace{-2mm}
%     \caption{Experimental results for different ground truth and workers' competence distribution where $\Delta_* = R(X, \mu) - R(\hat{\mu}_{EBE}(\alpha^*), \mu), \hat{\Delta}_*=2\alpha^*\Sigma_{j=1}^m Cov\Big(X_j, \frac{\hat{\sigma}^2(X_j-\bar{X})}{S^2}\Big) - (\alpha^*)^2E\Big(\frac{\hat{\sigma}^4}{S^2}\Big), \Delta = R(X, \mu) - R(\hat{\mu}_{EBE}(m-3), \mu)$. Where $\hat{\sigma}^2 = \frac{1}{n} \sum_{i=1}^n \frac{1}{m-1} \sum_{j=1}^m (X_{ij} - \bar{X}_j)^2$, and $\hat{X}_j = \frac{1}{n}\sum_{i=1}^n X_{ij}$ Each cell is a 10,000 samples average where the ground truth is fixed and the observations are sampled}\label{table:a*}
%     \end{table}
Proof is in Appendix \ref{proof:alpha_star}.
% Table \ref{table:a*} demonstrates how EBE yields a lower risk than BLUE and that $\alpha*$ further reduces that risk over different distributions, number of questions, and number of workers.
\subsection{Normal Model} \label{sec:normal}
Testing whether inequality (\ref{eq:cond}) holds could be a complicated task. Therefore, to get more intuition, in this subsection we reinstate the AWG model on our single aggregated worker. That is, we assume that for any question $j$, $X^A_j=\mu_j+\epsilon_j$, where the errors $\epsilon_j$ are sampled i.i.d. from $\mathcal{N}(0,\sigma^2)$. For a vector $\vec Y\in \mathbb{R}^m$ we denote by $\bar Y:=\frac{1}{m}\sum_jY_j$ and $S^2(Y):= \frac1{m-1}\sum_j(Y_j-\bar Y)^2$ its \emph{mean} and its \emph{sample variance}, respectively.
%\subsubsection{Single worker}
%Let $\vec{X} = (X_1,..,X_m)$ be the independent answers of a single worker to $m$ questions whose answers are $\vec{\mu} = (\mu_1,..,\mu_m)$ such that $X_j \sim \mathcal{N}(\mu_j, {\sigma}^2)$. 
%Here $\phi_{EBE}$ is the EBE as defined in Eq.~\eqref{eq:EBE} with  $\psi(X^A)=\hat{\sigma}^2$ replacing ${\sigma}^2$.\\\\
 %The following theorem calculates the risk of $\phi_{EBE}$. 

In addition, we assume that the variance estimator $\estn$ is a function of the \emph{aggregated observations} $\vec X^A$ (which may or may not be a sufficient statistic for $\sigma^2$), and that all of its directional derivatives exist.
\begin{theorem}\label{thm:single_worker_risk}
Under the Normal model, for $m>3$ 
\begin{multline}
\label{eq:delta_hat_sigma}
  \calR_{\vec{\mu}, \vec{\sigma}}(\ebAlg{\alg}{\estn}) = \calR_{\vec{\mu}, \vec{\sigma}}(\alg) \\+  \frac{(m-3)^2}{m-1}\Big(E_{\vec{\mu}, \vec{\sigma}}[\frac{(\estn(\vec X^A))^2}{S^2(\vec{X}^A)}]
  \\- 2{\sigma}^2\Big[E_{\vec{\mu}, \vec{\sigma}}[\frac{\estn(\vec X^A)}{S^2(\vec{X}^A)}] + E_{\vec{\mu}, \vec{\sigma}}\big[\frac{\Sigma_{j=1}^m\frac{d\psi(\vec X^A)}{dX^A_j}(X_j^A - \bar{X}^A)}{(m-3)S^2(\vec{X}^A)}\big]\Big]\Big)
\end{multline}
\end{theorem}
% \tbs{Should the $\vec{\sigma}$ be removed or replace with $\sigma$ or replaces with $\Bar{\sigma}$?}\rmr{no it's fine: $\vec\sigma$ is a parameter of the distribution $\calP$ used in the expectations and the risk.}
The proof is Appendix~\ref{proof:normalModel}\\
Theorem~\ref{thm:single_worker_risk} derives the explicit risk of the aggregated worker under general dependence structure of $\hat{\sigma}^2$ and $\vec{X}$. The expected reduction in risk when using empirical Bayes i.e, $\mathcal{R}_{\vec{\mu}, \vec{\sigma}}(\alg)- \mathcal{R}_{\vec{\mu}, \vec{\sigma}}(\ebAlg{\alg}{\estn})$  can be estimated from the observations, since if ${\sigma}^2$ is replaced with $\hat{\sigma}^2=\estn(\vec X^A)$ we get an expression which is exclusively dependent on the observations and thus, can be estimated.
%\rmr{how is that different from estimating the other expectations in the expression?}\tbs{If I got your question right then $\sigma$ is  a constant and does not have to be a function of the observations}\rmr{it is a constant in the same meaning that the other expectations are constants. In fact $\sigma^2 = E_{\vec\mu,\vec\sigma}[(X^A_j-\mu_j)^2]$ for any $j$}
Corollaries~\ref{cor:single_worker_cor1}-\ref{cor:meanAdjusted} extend the theorem and demonstrate different conditions which guarantee that EBE will have a lower risk than BLUE.
\begin{corollary}\label{cor:single_worker_cor1} Under the assumptions of Theorem~\ref{thm:single_worker_risk},
%\rmr{I'm not sure this corollary adds much}\tbs{it's a build up for the next points.. but yes I agree}\rmr{no harm in keeping it but if you need the space this has low priority}
 $$\mathcal{R}_{\vec{\mu}, \vec{\sigma}}(\ebAlg{\alg}{\estn}) < \mathcal{R}_{\vec{\mu}, \vec{\sigma}}(\alg) \ \forall \vec \mu \in \mathbb{R}^m, m>3$$ for any $\hat{\sigma}^2=\estn(\vec X^A)$ such that:
\begin{equation}\label{eq:single_worker_cor1}
    \frac{E_{\vec{\mu}, \vec{\sigma}}[\frac{\hat{\sigma}^4}{S^2}]}{E_{\vec{\mu}, \vec{\sigma}}[\frac{\hat{\sigma}^2}{S^2(\vec{X}^A)}] + E_{\vec{\mu}, \vec{\sigma}}\big[\frac{\Sigma_{j=1}^m\frac{d\estn(\vec X^A)}{dX_j^A}(X_j^A - \bar{X}^A)}{(m-3)S^2(\vec{X}^A)}\big]} < 2{\sigma}^2
\end{equation}
\end{corollary}

We promised that under the Normal model we would get more intuition, but Condition~\eqref{eq:single_worker_cor1} is not quite there yet. Note however that if $\estn$ is a constant function (i.e., $\hat\sigma^2$ is guessed or estimated not from the data), then a whole chunk of the expression disappears. We next show that this still occurs under a less restrictive assumption.

\paragraph{Mean-adjusted estimators}
If we use a reasonable variance estimator $\estn$, we would expect a lower estimation as observations are closer to their mean. 
%When estimating the variance of a single worker from $\vec{X}$ a reasonable estimator for the variance becomes smaller when any entry of $\vec{X}$ is getting closer to $\bar{X}$ and vise versa. For this case we provide the following corollary.

\begin{definition}
An estimator $\estn(\vec X)$ is \emph{mean-adjusted} if  for each coordinate $j$ such that $X_j \leq \bar{X}$ (respectively, $X_j > \bar{X}$), we have $\frac{d}{dX_j}\psi(\vec X) \leq 0$ (respectively, $\frac{d}{dX_j}\psi(\vec X) \geq 0$).
\end{definition}
It is not hard to find estimators that are mean-adjusted, for example $\psi_S(\vec X):=S^2(\vec X)c$ for any constant $c\geq 0$.
\begin{corollary}\label{cor:meanAdjusted}
Under the Normal model,  if $\estn$ is mean-adjusted then 
\begin{align*}
    &\mathcal{R}_{\vec{\mu}, \vec{\sigma}}(\ebAlg{\alg}{\estn}) < \mathcal{R}_{\vec{\mu}, \vec{\sigma}}(\alg) 
    \\ &+ 
    \frac{(m-3)^2}{m-1}(E_{\vec{\mu}, \vec{\sigma}}[\frac{(\estn(\vec X^A))^2}{S^2(\vec{X}^A)}] - 2{\sigma}^2E_{\vec{\mu}, \vec{\sigma}}[\frac{\estn(\vec X^A)}{S^2(\vec{X}^A)}])
\end{align*}
And hence, $\mathcal{R}_{\vec{\mu}, \vec{\sigma}}(\ebAlg{\alg}{\estn}) < \mathcal{R}_{\vec{\mu}, \vec{\sigma}}(\alg)$ for all $\vec \mu \in \mathbb{R}^m$ and  $m>3$ for any $\hat{\sigma}$ which satisfies:
\begin{equation}\label{eq:mean_adjusted}
\frac{E_{\vec{\mu}, \vec{\sigma}}[\frac{(\estn(\vec X^A))^2}{S^2(\vec{X}^A)}]}{E_{\vec{\mu}, \vec{\sigma}}[\frac{\estn(\vec X^A)}{S^2(\vec{X}^A)}]} < 2{\sigma}^2   
\end{equation}
\end{corollary}

Now Condition~\eqref{eq:mean_adjusted} is simple enough to provide some intuition. To see this even better we consider two special cases of estimators:
\begin{enumerate}
    \item The first special case is when $\hat\sigma^2=\estn(\vec X)$ is a constant. Then  Condition~\eqref{eq:mean_adjusted} simplifies to $\hat\sigma^2 < 2\sigma^2$: for a `correct guess' $\hat\sigma^2=\sigma^2$ we get the maximal improvement, which becomes weaker as $\hat\sigma^2$ drifts towards $0$ or $2\sigma^2$.
    \item The second special case is $\estn(\vec X)= S^2(\vec{X})$.  Condition~\eqref{eq:mean_adjusted} then simplifies to
$$E_{\vec{\mu}, \vec{\sigma}}[S^2(\vec{X}^A)] < 2{\sigma}^2.$$
Since $E[S^2(\vec X^A)]$ is itself roughly proportional to the squared error $\sigma^2$ plus $S^2(\vec \mu)$, we get a valuable indication that Empirical Bayes is expected to perform better on more uniform sets of questions (i.e. whose true answers do not vary substantially).\rmr{this is actually true even in the general model, and we can always satisfy the condition by setting $\estn(\XX)= \|\vec X-\bar X\vec a\|/k$ for sufficiently large $k$. see test.tex.}
\end{enumerate}
  
%   \rmr{This is the third time this comment appears so I think we can remove it:}
% Corollaries~\ref{cor:single_worker_cor1}-\ref{cor:meanAdjusted} theoretically suggest a way of estimating from the data whether EBE would dominate BLUE together with replacing $\sigma^2$ with $\hat{\sigma}^2$ on the right side of the inequalities. \rmr{the following sentence unclear (and I think unnecessary):} Specifically, if the derivatives of $\hat{\sigma}^2$ w.r.t $\vec{X}$ can be explicitly computed/estimated then Corollary~\ref{cor:single_worker_cor1} can be used, for the case where those derivatives cannot be used but $\hat{\sigma}^2$ behaves "reasonably, i.e. if we take a coordinate $X_j$ closer to the mean of $\vec{X}$ then the resulted $\hat{\sigma}^2$ should be smaller or equal.
\par We provide the proof of corollary \ref{cor:meanAdjusted} in Appendix~\ref{proof:meanAdjusted}. We show more corollaries in Appendix~\ref{apx:additional_corollaries}, and in Appendix~\ref{apx:deterministic_est} we deal with the special case of single or multiple workers where $\hat{\sigma}^2$ is a constant.

\section{Empirical Evaluation}\label{sec:emp}
In this section, we show experimental results over different datasets and algorithms. We evaluate EBE combined with various truth-discovery algorithms.

That is, we run our $\ebAlg{\alg}{\estn}$ algorithm, where the baseline truth-discovery algorithm $\alg$ varies (see below). For the variance estimator $\estn$ we use the following heuristic, which estimates the variance of each worker using $\vec X^A$ as a proxy of the truth, and then takes the average:
$$\estn_H(\XX): = \frac{1}{n} \sum_{i=1}^n \frac{1}{m-1} \sum_{j=1}^m (X_{ij} - \bar{X}^A_j)^2. $$

% \par \rmr{This paragraph seems redundant:}Denote $X_A=(X_A^1,\dots,X_A^m)$ as the truth estimation output of algorithm A, where $X_A^j = \sum_{i=1}^n w_A^iX_{ij}$ and $w_A=(w_A^1,\dots,w_A^n)$ are the weights assigned by the algorithm. Here A can be any TD algorithm. %We then apply both definition \ref{EBE} and definition \ref{stein_estimator} by setting $X=X_A$ using a simple estimation of the aggregated worker $\hat{\sigma}^2 = \frac{1}{n} \sum_{i=1}^n \frac{1}{m-1} \sum_{j=1}^m (X_{ij} - \bar{X}^A_j)^2$ %\David{$\hat{\sigma}^2 = \frac{1}{n} \sum_{i=1}^n \frac{1}{m-1} \sum_{j=1}^m (X_{ij} - \hat{X}_j)^2$??}
%where $\hat{X}_j = \frac{1}{n}\sum_{i=1}^n X_{ij}$. We then compute the risk ratio of EBE over BLUE to test whether using EBE yields a lower risk when applied over $A$'s output.

\newpar{Algorithms} 
We use the following truth-discovery algorithms from the literature: GTM~\cite{GTM}, CATD~\cite{CATD}, and KDEm~\cite{KDEm}, IPTD~\cite{meir2021generaldomain}, DTD~\cite{grofman1983thirteen} and, CRH~\cite{CRH}. We let each of the aforementioned algorithms up to 14 iterations to converge. 
%\rmr{I would remove DTD- Grofman only defined something similar for binary domain, and it anyways coincides with PTD}\tbs{I don't think I have enough time for making new plots without DTD}\rmr{you can just rename it to PTD}\tbs{But PTD is not iterative.. otherwise I just have two IPTD, how about instead, I can just write a sentence about how DTD was used}
\newpar{Datasets}
We used datasets from the following papers: Buildings~\cite{meir2021generaldomain} where 208 workers answered 25 questions; Triangles1-Triangles2~\cite{hart2018statistical} where 50 workers answered 300 questions; Emotions1-Emotions4~\cite{snow2008cheap} where 10 workers answered 200 questions;
In addition we have generated synthetic datasets using the AWG model, such that $X_{ij} \sim N(\mu_j, \sigma_i^2)$, where $\sigma_i^2 \sim N(1, 0.5)$. The distribution of the ground truth $\mu_j$ appears on top of each figure.
%\David{Maybe you should say more about the datasets. How many workers? How many tasks?} \rmr{generated datasets are not `from' any particular paper. You just need to describe the distribution. }

\newpar{Evaluation}
To investigate whether EBE can lower the risk of the above TD algorithms, we sample a subset of workers and questions, run each algorithm and compute the \emph{Improvement Ratio}:
$$IR := \frac{\mathcal{R}(\ebAlg{\alg}{\estn_H})}{\mathcal{R}(\alg)},$$ 
where the (empirical) risk is calculated by taking the average over 1000 samples of $n$ workers and $m$ questions from the dataset. 

An \emph{Improvement Ratio} (IR) $<1$ indicates that Empirical Bayes improves the baseline algorithm $\alg$ on this dataset. 

\begin{figure}
    \centering
    \includegraphics[width=8cm, height=10cm]{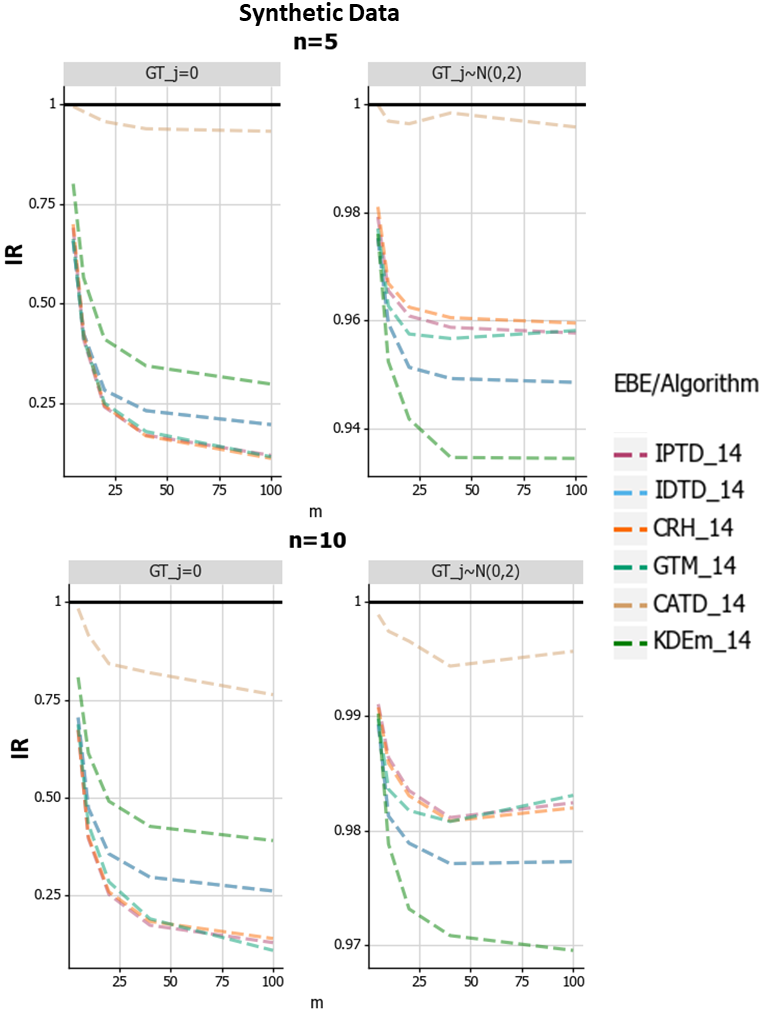}
    \caption{IR over synthetic data sets, $n$ is the number of workers and $m$ is the number of questions the left plots are under a constant GT, and the right plots are under a random GT}
    \label{fig:synthetic}
\end{figure}

 %\David{I think that you should look at ratios, instead of differences and discuss improvements (or lack of improvements) in terms of percents. Also, think on graphic ways to show parts of Tables 3 and 4. Also, as I understand, there is no dataset where EBE really helps. This is not so good. Maybe you can divide the detests to more homogeneous parts and apply EBE for each part separately?  }
\newpar{Results}
In the synthetic datasets (Figure~\ref{fig:synthetic}, left) we see that when the ground truth (GT) is constant, EB significantly improves all algorithms, lowering the risk by a factor of 1\% - 90\%. When the variance of the ground truth is higher (right figures) the IR is closer to $1$.

\begin{figure}
    \centering
    \includegraphics[width=.99\linewidth]{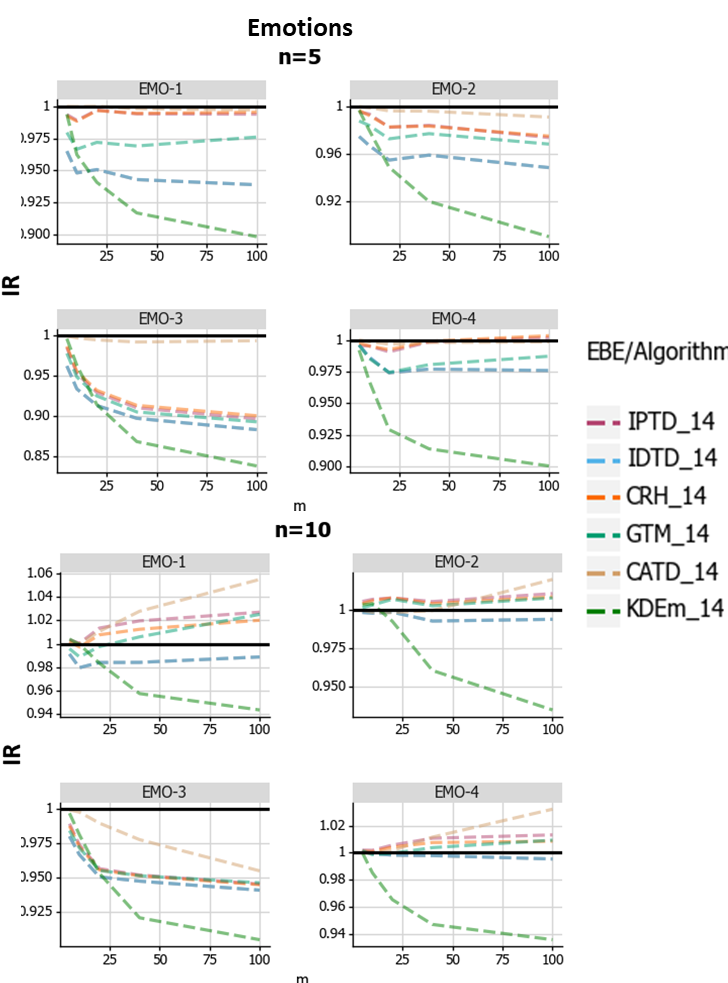}
        \caption{$Risk$ $Ratio$ for the Emotions datasets} 
    \label{fig:emotions}
\end{figure}

\begin{figure}
    \centering
    \includegraphics[width=1\linewidth]{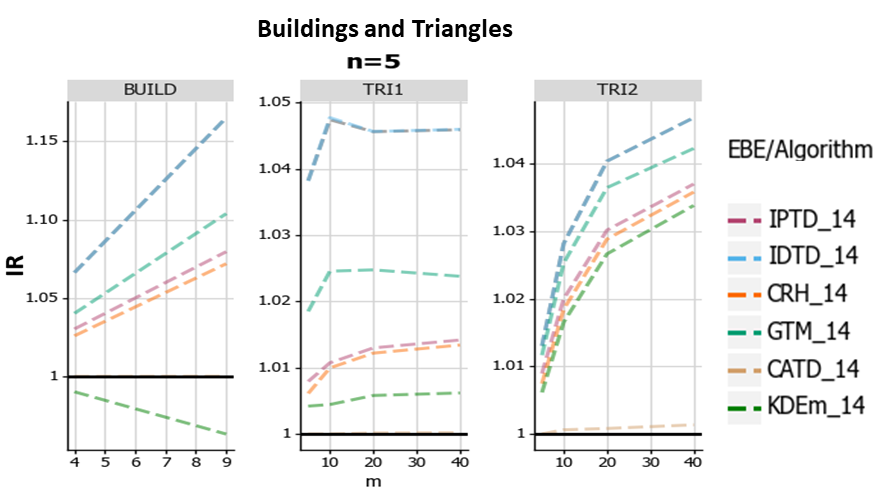}
     \caption{$Risk$ $Ratio$ for the Buildings and Triangles datasets} 
    \label{fig:build_tri}
\end{figure}

In real world datasets, results are mixed. In the Emotions datasets (Figure~\ref{fig:emotions}) there is an improvement, especially when $n$ is low. 
Figure~\ref{fig:build_tri} shows real-world datasets where the ground truth is highly variable, compared to the noise. This high variance  causes EB to fail. 

\begin{figure}
    \centering
    \includegraphics[width=6cm, height=4cm]{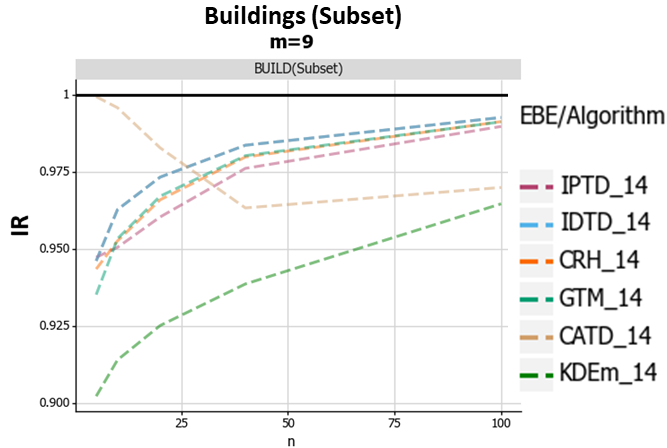}
    \caption{IR of a subset of the Buildings dataset reducing the underlying GT variance from 51567 to 633}
    \label{fig:subset_build}
\end{figure}

\begin{figure}\hspace{-1.25cm}
    \centering
    \includegraphics[width=7cm, height=3cm]{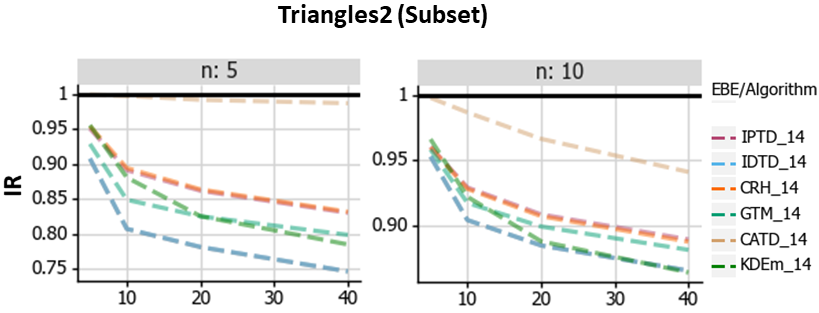}
    \caption{IR of a subset of Triangles2 dataset thus reducing the underlying GT variance from 65362 to 195}
    \label{fig:subset_tri}
\end{figure}

However recall that we recommended based on the discussion following Cor.~\ref{cor:meanAdjusted} working with a more `uniform' sets of questions.
To test this point in practice, we partitioned the questions and considered `uniform' subsets where the variance of the ground truth is low.
Indeed, Fig.~\ref{fig:subset_build} and Fig.~\ref{fig:subset_tri} show that on the low-variance datasets, Empirical Bayes improves the outcome and reduces the error.

To conclude, Empirical Bayes is particularly effective when there are few workers and low variance of the ground truth, and this applies regardless of the baseline truth discovery algorithm in use. % when EBE is applied to questions which their underlying GT has relatively low variance, applying EBE shows great advantage, especially when there is a small number of workers.

\section{Related work}\label{sec:relatedwork}
As we stated in the introduction, a large part of the truth discovery literature deals with estimating workers' competence. Of those that deal with real-valued data, most make iterative estimations of the ground truth and the competence, and differ in how they implement the steps. For example,  \citet{meir2021generaldomain} show that workers' average distance between answers from the other workers' answers can estimate their competence; \citet{CATD} weigh workers' responses proportionally to the upper confidence interval limit of their estimated variance; \citet{CRH} weigh workers' responses by a convex optimization framework which, ``minimizes the weighted deviation from the truths to the multi-source input".  A different approach taken by \citet{KDEm} is weighing the responses by the weights which minimizes the kernel density estimation applied to each question separately. Often algorithms use the BLUE estimator with the estimated competences instead of the true workers' variance. 

Surveys (such as  \cite{survey}) show that there is no single state-of-the-art.
Some algorithms work better than others on specific domains and worse on different domains, %each of them outputs their estimations for the true answers (the actual buildings’ heights in our example) via estimating the workers’ competence; they either output the mean answer of the k-most competent workers, or they use BLUE. This research paper attempts to improve each of the described algorithms’ (and others) outputs by adding one more step and using their estimations of workers’ competence.
This highlights the importance of methods that are not algorithm-specific.

A closely related work to ours is of \citet{GTM}, where a Bayesian approach is taken. The authors assume prior distributions over the ground truth (Normal distribution), and workers' competence (Beta distribution). Then, an Expectation Maximization (EM) approach is taken for the estimation of workers' competence. The algorithm's output is the posterior mean of the ground truth which incorporates chosen hyper-parameters (prior knowledge) and the estimated competence. %Although in this paper we also assume a normal ground truth prior\rmr{ where do we assume that??}\tbs{That's what Bayesian here, to derive EBE you go through assuming a prior etc.}, 
We estimate the posterior differently following \cite{stein1956inadmissibility}, we do not incorporate any hyper-parameters. %, we assume the estimation of competences are either known or estimated,
Most importantly, our results rely on theoretical foundations.

Other truth-discovery algorithms that deal with binary or categorical labels are outside the scope of this work. 
%\par Many algorithms deal with continuous data and estimate workers' competence (One TD algorithms survey is \cite{survey}). \citet{grofman1983thirteen} (Theorem VII) show a fundamental approach to estimating a worker’s competence by their agreement with the majority choices. 
\section{Conclusion}
We showed that when workers' competences are known, the Empirical Bayes approach is \emph{always} a better choice (when there are more than $3$ questions), and improves  \emph{any} TD algorithm that does not have access to workers' competences, for an appropriate variance estimator.

We demonstrated both in theory and in practice that the potential improvement of EBE depends on the uniformity of the set of questions (i.e. it works better when applied to questions whose answers are similar). On the other hand performance also improves when applied to more questions, thus we have an inherent tradeoff between grouping many questions together, or separate them to smaller chunks of `similar' questions.

Future work might consider how to integrate this into the algorithm, by appropriately partitioning the questions in a way that maximizes the benefit of  EBE.

\section{Acknowledgements}
This research was supported by THE ISRAEL SCIENCE FOUNDATION
(grant No. 2539/20).

\clearpage
\begin{small}
 \bibliography{eb.bib}
 \end{small}
\bibliographystyle{abbrvnat}
\clearpage
\appendix
\onecolumn
\begin{small}

\section{Deriving The Empirical Bayes Estimator (EBE)}\label{proof:EBEderive}
% \rmr{maybe add a few words on why you derive this part again if it is supposed to be known. Also to make it clear which parts are new.}
In Appendix A we present known results as a background for the Empirical Bayes approach.\\
On our path to deriving an Empirical Bayes Estimator (EBE) for $\vec{\mu}$ we first need to derive the appropriate Bayes estimator. Recall $\mathcal{L}(\alg{(X)}, \vec{\mu})$ is a loss square euclidean loss function, i.e, $\mathcal{L}(\alg{(X)}, \vec{\mu}) = \|\vec{\mu} - \alg(X)\|_{l2}^2$. 
\newpar{Bayesian Estimator}
 The Bayesian Estimator minimizes the Bayes risk which we will soon present, but first, we introduce the notion of the Prior distribution.
\newpar{Prior probability distribution}
The Bayesian framework assumes a known prior, this will later we will relax the known-prior-assumption and estimate its parameters. First, we assume $\vec{\mu} = (\mu_1,..,\mu_m)$ is normally distributed, that is, we assume
\begin{equation*}\label{prior0}
  \mu_j \sim \mathcal{N}_0 = \mathcal{N}(\mu_0, \sigma_0^2)  
\end{equation*}
For example we can think of a group of workers answering "what are the weights of the people is these images?", thus, the underlying ground truth are weights of people which are close to a normal distribution.
\newpar{Observations $\vec{X} = (X_1,\dots,X_m)$}
Assuming a prior distribution over $\vec{\mu}$, (notice the difference from the preliminaries where $X_j \sim \mathcal{N}(\mu_j, \sigma^2)$), we now denote the conditional distribution $X_j| \mu_j \sim \mathcal{N}(\mu_j, \sigma^2)$ 
\newpar{Posterior probability distribution $\mu_j | X_j$}
can be viewed as an update for the assumed prior distribution after viewing the observations. It is well known that the Bayesian Estimator (which we will later show that it is better than the BLUE estimator shown in Theorem~\ref{thm:Aitkin}) for the square loss function is the posterior's mean, that is, $E_{\mu\sim N_0, \sigma}(\mu_j|X_j)= \min_{\alg(X_j)}{E_{\mu\sim N_0, \sigma}(\alg(X_j)-\mu_j)^2}$ where the latter expectation is under the Bayesian framework, that is $E_{\mu\sim N_0, \sigma}(\alg(X_j)-\mu_j)^2 = \int_\mu E_{\vec{\mu}, \vec{\sigma}}(\alg(X_j)-\mu)^2d\mu$,  hence, to estimate $\mu_j$ via a Bayesian Estimator we first need to calculate the posterior mean.
\begin{theorem}\label{theoremPost}
The posterior distribution is $\mu_j|X_j \sim \mathcal{N}(\tilde{\mu}_j, \tilde{\sigma}^2)$, 
\begin{equation*}\label{posterior}
\tilde{\mu}_j = X_j * \frac{\sigma^2_{0}}{\sigma^2_{0} + \sigma^2} + \mu_0 * \frac{\sigma^2}{\sigma^2_{0} + \sigma^2}, \ \tilde{\sigma}^2 = \frac{\sigma^2_{0}\sigma^2}{\sigma^2_{0} + \sigma^2}
\end{equation*}
\end{theorem}
\begin{proof}
\begin{align*}
    \mathcal{P}(\mu_j|X_j) &= \frac{\mathcal{P}(X_j|\mu_j)\mathcal{P}(\mu_j)}{\mathcal{P}(X_j)} \propto \mathcal{P}(X_j|\mu_j)\mathcal{P}(\mu_j) \\
    &\propto e^{\frac{\frac{(X_j-\mu_j)^2}{\sigma^2} + \frac{(\mu_j-\mu_0)^2}{\sigma_0^2}}{-2}} \propto e^{\frac{\mu_j^2(\frac{1}{\sigma^2} + \frac{1}{\sigma_0^2}) -2\mu_j(\frac{X_j}{\sigma^2} + \frac{\mu_0}{\sigma_0^2})}{-2}} \propto e^{\frac{(\frac{1}{\sigma^2} + \frac{1}{\sigma_0^2})(\mu_j-\tilde{\mu})^2}{-2}}
\end{align*}
Where the last proportional-to follows from completing-the-square technique and $$\tilde{\mu}_j = \frac{X_j\sigma_0^2+\mu_0\sigma^2}{\sigma_0^2 + \sigma^2}$$
\end{proof}
\subsection{The Empirical Bayes Estimator (EBE)}
We now relax the previous known-prior-assumption, this subsection is the empirical part in Empirical Bayes. We will show how to estimate the posterior mean using the observed data.\\
First we will derive the marginal distribution of $X_j$, that is, we evaluate the following expression: $$\mathcal{P}_{\mu_j\sim \mathcal{N}_0, \sigma}(X_j)=\int_{\mu_j}{\mathcal{P}_{\mu_j, \sigma}(X_j|\mu_j)\mathcal{P}(\mu_j)}d\mu_j$$
But we already saw on (\ref{theoremPost}) that $\mathcal{P}_{\mu_j, \sigma}(X_j|\mu_j)\mathcal{P}(\mu_j)$ is normally distributed, thus what is left is to calculate its mean and variance.\\
By law of total expectation
\begin{equation*}
    E_{\mu_j\sim N_0, \sigma}(X_j) = E_{\mu_j \sim N_0, \sigma}(E_{\mu_j, \sigma}(X_j|\mu_j)) = E_{\mu_j \sim N_0, \sigma}(\mu_j) = \mu_0
\end{equation*}
By law of total variance
\begin{equation*}
    Var_{\mu_j \sim N_0, \sigma}(X_j) = E_{\mu_j \sim N_0, \sigma}(Var_{\mu_j, \sigma}(X_j|\mu_j)) + Var_{\mu_j \sim N_0, \sigma}(E_{\mu_j, \sigma}(X_j|\mu_j)) = \sigma^2 + \sigma_0^2
\end{equation*}
\newpar{Empirical Bayes Estimator}
Similarly to the work \emph{An introduction to Empirical Bayes Data Analysis} [George Casella, 1985], we can construct an estimate for $\tilde{\mu}_j$ at (\ref{posterior}), denote $\bar{X} = \frac{1}{m} \sum_{j=1}^m X_j$

\begin{proposition}\label{propChi}
$E_{\mu\sim N_0, \sigma}[\frac{\sigma^2 + \sigma_0^2}{\sum_{j=1}^m(X_j-\bar{X})^2}] = \frac{1}{m-3}$
\end{proposition}
\begin{proof}
Notice that the expectation is over the marginal of $X_j$ and thus, the $X_j$'s are i.i.d which means the expectation is of an inverse-chi-squared distribution with $m-1$ degrees of freedom. 
\end{proof}

An immediate corollaries of \ref{propChi} are:
\begin{corollary}\label{B_Hat}
$E_{\mu\sim N_0, \sigma}[\frac{(m-3)\sigma^2}{\sum_{j=1}^m(X_j-\bar{X})^2}] = \frac{\sigma^2}{\sigma^2 + \sigma_0^2}$
\end{corollary}

\begin{corollary}An Empirical Bayes Estimator is
\begin{equation*}\label{casella}
\EB(\vec{X}, \sigma) = \frac{(m-3)\sigma^2}{\sum_{j=1}^m (X_j-\bar{X})^2}\bar{X}\boldsymbol{1}  + \Big[1 - \frac{(m-3)\sigma^2}{\sum_{j=1}^m (X_j -\bar{X})^2}\Big]\vec{X}  
\end{equation*}

\end{corollary}

\section{Proofs}
\subsection{Theorem \ref{cassellaBound} (Casella, 1985's Theorem)}\label{proof:casella}

In the AWG model with a single worker and $m>3$ questions, 
\begin{equation}
    \mathcal{R}_{\vec{\mu},{\sigma}}(\EBv) < \mathcal{R}_{\vec{\mu},\sigma}(\estm_I) \ \text{ for all }\vec{\mu} \in \mathbb{R}^m, {\sigma} \in \mathbb{R}^+.
\end{equation}    

\begin{proof}
Denote $\hat{B} = \frac{(m-3)\sigma^2}{S^2(\vec{X})}$
\begin{align*}
   \mathcal R_{\vec{\mu}, \sigma}(\EB) &= E_{\vec{\mu}, \sigma}[\lVert \bar{X} + (1-\hat{B})(X - \bar{X}) - \mu \rVert^2] \\
    &=\Sigma_{j=1}^m E_{\vec{\mu}, \sigma}[(\bar{X} + (1-\hat{B})(X_j - \bar{X}) - \mu_j)^2]\\
    &= \Sigma_{j=1}^m E_{\vec{\mu}, \sigma}[(X_j - \mu_j - \hat{B}(X_j - \bar{X}))^2]\\
    &= \Sigma_{j=1}^m E_{\vec{\mu}, \sigma}[(X_j - \mu_j)^2] - 2E_{\vec{\mu}, \sigma}[ \hat{B}(X_j - \mu_j)(X_j - \bar{X})] + E_{\vec{\mu}, \sigma}[(\hat{B}(X_j - \bar{X}))^2]\\
    &= \mathcal{R}_{\vec{\mu},\sigma}(\estm_I) - [\Sigma_{j=1}^m2E_{\vec{\mu}, \sigma}[ \hat{B}(X_j - \mu_j)(X_j - \bar{X})]] + \sigma^4(m-3)^2E_{\vec{\mu}, \sigma}[\frac{1}{S^2(\vec{X})}]
\end{align*}
\begin{lemma} Stein's Lemma: \\
Let $X\sim N(\mu, \sigma^2)$ and Let $g(X)$ be a function for which $E_{\vec{\mu}, \sigma}[g(X)(X-\mu)]$ and $E_{\vec{\mu}, \sigma}[\frac{d}{dx}g(X)]$ both exist, then $E_{\vec{\mu}, \sigma}[g(X)(X-\mu)] = \sigma^2E_{\vec{\mu}, \sigma}[\frac{d}{dx}g(X)]$
\end{lemma}
We now focus on the mixed term:\\
Denote $g(X_j) = \frac{X_j - \bar{X}}{S^2(\vec{X})}$\\
\begin{align*}
    E_{\vec{\mu}, \sigma}[ \hat{B}(X_j - \mu_j)(X_j - \bar{X})] &= \sigma^2(m-3)E_{\vec{\mu}, \sigma}[\frac{X_j-\bar{X}}{S^2(\vec{X})}(X_j-\mu_j)]\\
    &= \sigma^2(m-3)E_{\vec{\mu}, \sigma}[E_{\vec{\mu}, \sigma}[\frac{X_j-\bar{X}}{S^2(\vec{X})}(X_j-\mu_j)|X_1,\dots,X_{j-1}, X_{j+1},\dots, X_m]]\\
    &=\sigma^2(m-3)E_{\vec{\mu}, \sigma}[E_{\vec{\mu}, \sigma}[g(X_j)(X_j-\mu_j)|X_1,\dots,X_{j-1}, X_{j+1},\dots, X_m]]\\
    &=_{stein}\sigma^4(m-3)E_{\vec{\mu}, \sigma}[E_{\vec{\mu}, \sigma}[\frac{d}{dX_j}g(X_j)|X_1,\dots,X_{j-1}, X_{j+1},\dots, X_m]]\\
    &= \sigma^4(m-3)E_{\mu,
    \sigma}[\frac{d}{dX_j}g(X_j)]\\
    \frac{d}{dX_j}g(X_j) &= \frac{(1 - \frac{1}{m})S^2(\vec{X}) -2(X_j-\bar{X})^2}{\big (S^2(\vec{X})\big )^2}
\end{align*}
The mixed term can now be rewritten as follows:
\begin{align*}
    -\Sigma_{j=1}^m2E_{\vec{\mu}, \sigma}[ \hat{B}(X_j - \mu_j)(X_j - \bar{X})] &= -2\sigma^4(m-3)E_{\vec{\mu}, \sigma}[\frac{(1 - \frac{1}{m})mS^2(\vec{X}) -2S^2(\vec{X})}{\big ( S^2(\vec{X}) \big )^2}]\\
    &= -2\sigma^4(m-3)^2E_{\vec{\mu}, \sigma}[\frac{1}{S^2(\vec{X})}]
\end{align*}
Summing up everything:
\begin{align*}
 \mathcal R_{\vec{\mu}, \sigma}(\EB) &=\\
 &= \mathcal{R}_{\vec{\mu},\sigma}(\estm_I) -2\sigma^4(m-3)^2E_{\vec{\mu}, \sigma}[\frac{1}{S^2(\vec{X})}] + \sigma^4(m-3)^2E_{\vec{\mu}, \sigma}[\frac{1}{S^2(\vec{X})}]\\
 &= \mathcal{R}_{\vec{\mu},\sigma}(\estm_I) - \sigma^4(m-3)^2E_{\vec{\mu}, \sigma}[\frac{1}{S^2(\vec{X})}] \ \forall m > 3
\end{align*}

\end{proof}
\subsection{Proposition ~\ref{prop:suff_stat}}\label{suff_statistic}
$\algv_B(\boldsymbol{X})$ is  a sufficient statistic for $\vec{\mu} = (\mu_1,..,\mu_m)$ under $\mathcal{P}_{\vec{\mu}, \vec{\sigma}}$.
\begin{proof}Since the regularity conditions hold for a multiplication of independent normal distributions, to prove sufficiency by the Fisher-Neyman factorization theorem, we only need to show that the model can be represented as a multiplication of a function $\tilde{h}$ of the observations and a function $\tilde{g}$ of a sufficient statistic and the unknown parameter as follows:
\begin{align*}
    \mathcal{P}_{(\mu_1,..,\mu_m, \sigma_1, \dots, \sigma_n)}(X_{11},..,X_{nm}) &= \Pi_{j=1}^m\Pi_{i=1}^n\mathcal{P}_{\mu_j, \sigma_i}(X_{ij})\\
    &= \Pi_{j=1}^m\Pi_{i=1}^n\frac{1}{\sigma_i\sqrt{2\pi}}e^{-\frac{1}{2\sigma_i^2}(X_{ij}-\mu_j)^2}\\ 
    &= \Pi_{j=1}^m[C \cdot e^{-\frac{1}{2}\sum_{i=1}^n\frac{X_{ij}^2}{\sigma_i^2}}]*[e^{-\frac{1}{2}\mu_j^2\sum_{i=1}^n\frac{1}{\sigma_i^2} + \mu_j \sum_{i=1}^n\frac{X_{ij}}{\sigma_i^2}}]\\ 
    &= \Pi_{j=1}^m h(X_{1j},..,X_{nj})g(\delta_j(X), \mu_j)\\
    &=\tilde{h}(X_{11},..,X_{nm})\tilde{g}(\delta(X), \mu)
\end{align*}
where C is a constant.
\end{proof}
\subsection{Theorem \ref{general_model}}\label{proof:general_model}
For any unbiased algorithm $\alg$, and $m>3$,\footnote{Since in this subsection we do not assume that the distribution of $\XX$ follows the AWG model, we do not need a parameter for the individual competence. Other than that, all definitions remain the same.}

$$\calR_{\vec\mu}(\ebAlg{\alg}{\estn}) <\calR_{\vec\mu}(\alg) \text{ for all }\vec{\mu} \in \mathbb{R}^m $$
if and only if
\begin{equation}
  2(m-3)\Sigma_{j=1}^m Cov\Big(X_j^A, \frac{\estn(\XX)(X_j^A-\bar{X}^A)}{\|\vec{X}^A -\bar{X}^A{\vec{1}}\|^2}\Big) 
  -(m-3)^2E_{\vec\mu}\Big(\frac{(\estn(\XX))^2}{\|\vec{X}^A -\bar{X}^A{\vec{1}}\|^2}\Big) >  0.
\end{equation}
For convenience we denote $\vec{X} = \vec{X}^A$
\begin{proof}
\begin{align*}
\calR_{\vec\mu}(\ebAlg{\alg}{\estn}) &= E_{\vec{\mu}}[\lVert \bar{X} + (1-\frac{\alpha\hat{\sigma}^2}{S^2(\vec{X})})(X - \bar{X}) - \mu \rVert^2] \\
    &= E_{\vec{\mu}}[\|(X - \mu)\|^2] - 2E_{\vec{\mu}}[ \frac{\alpha\hat{\sigma}^2}{S^2(\vec{X})}(X - \mu)^T(X - \bar{X})] + E_{\vec{\mu}}[(\|\frac{\alpha\hat{\sigma}^2}{S^2(\vec{X})}(X - \bar{X})\|^2]\\
    &= \calR_{\vec\mu}(\alg) -2\alpha E_{\vec{\mu}}[ \frac{\hat{\sigma}^2}{S^2(\vec{X})}(X - \mu)^T(X - \bar{X})] + \alpha^2E_{\vec{\mu}}[\frac{\hat{\sigma}^4}{S^2(\vec{X})}]\\
    &\text{Focusing on the mixed term:}\\
    E_{\vec{\mu}}[ \frac{\hat{\sigma}^2}{S^2(\vec{X})}(X - \mu)^T(X - \bar{X})] &= \sum_{i=1}^mE_{\vec{\mu}}[ \frac{\hat{\sigma}^2}{S^2(\vec{X})}(X_j - \mu_j)(X_j - \bar{X})]\\
    &=_{E_{\vec{\mu}}(X_j)=\mu_j} \sum_{i=1}^m Cov\Big(X_j - \mu_j, \frac{\hat{\sigma}^2(X_j - \bar{X})}{S^2(\vec{X})}\Big)\\
    &= \sum_{i=1}^m Cov\Big(X_j, \frac{\hat{\sigma}^2(X_j - \bar{X})}{S^2(\vec{X})}\Big)\\
    \text{Therefore we get:}\\
    \calR_{\vec\mu}(\alg) - \calR_{\vec\mu}(\ebAlg{\alg}{\estn}) &=  2\alpha\sum_{i=1}^m Cov\Big(X_j, \frac{\hat{\sigma}^2(X_j - \bar{X})}{S^2(\vec{X})}\Big) - \alpha^2E_{\vec{\mu}}[\frac{\hat{\sigma}^4}{S^2(\vec{X})}]\\
\end{align*}
\end{proof}
\subsection{Proposition \ref{prop:alpha_star}}\label{proof:alpha_star}
Denote $\psi(\XX) = \hat{\sigma}^2$\\
Choosing $\alpha^* = \frac{\Sigma_{j=1}^m Cov\Big(X_j^A, \frac{\hat{\sigma}^2(X_j-\bar{X})}{S^2(\vec{X}^A)}\Big)}{E_{\vec{\mu}}\Big(\frac{\hat{\sigma}^4}{S^2(\vec{X}^A)}\Big)}$ minimizes $\calR_{\vec{\mu}}(\ebAlg{\alg}{\estn})$.

\begin{proof}
Denote  $C := Cov\Big(X_j^A, \frac{\hat{\sigma}^2(X_j^A-\bar{X}^A)}{S^2(\vec{X}^A)}\Big)$ and  $E := E_{\vec{\mu}}\Big(\frac{\hat{\sigma}^4}{S^2(\vec{X}^A)}\Big)$. Then from Theorem~\ref{general_model} We wish to maximize the parabola $2\alpha C -\alpha^2E$ which can be easily shown to maximized at $\alpha^* = \frac{C}{E}$.
\end{proof}
\subsection{Theorem \ref{thm:single_worker_risk}}\label{proof:normalModel}
Denote $\psi(\vec{X})=\hat{\sigma}^2$ as an estimator of $\sigma^2$.

\begin{multline}
  \calR_{\vec{\mu}, \vec{\sigma}}(\ebAlg{\alg}{\estn}) = \calR_{\vec{\mu}, \vec{\sigma}}(\alg) +  \frac{(m-3)^2}{m-1}\Big(E_{\vec{\mu}, \vec{\sigma}}[\frac{(\estn(\vec X^A))^2}{S^2(\vec{X}^A)}]
  - 2{\sigma}^2\Big[E_{\vec{\mu}, \vec{\sigma}}[\frac{\estn(\vec X^A)}{S^2(\vec{X}^A)}] + E_{\vec{\mu}, \vec{\sigma}}\big[\frac{\Sigma_{j=1}^m\frac{d\psi(\vec X^A)}{dX^A_j}(X_j^A - \bar{X}^A)}{(m-3)S^2(\vec{X}^A)}\big]\Big]\Big)
\end{multline}

For convenience we denote $\vec{X} = \vec{X}^A$
\begin{proof}
Denote $g(X) = \frac{\hat{\sigma}^2(X_j - \bar{X})}{S^2(\vec{X})}$
\begin{align*}
    \calR_{\vec{\mu}, \vec{\sigma}}(\ebAlg{\alg}{\estn}) &= E_{\vec{\mu}, \vec{\sigma}}[\lVert \bar{X} + (1-\hat{B})(X - \bar{X}) - \mu \rVert^2] \\
    &= \Sigma_{j=1}^m E_{\vec{\mu}, \vec{\sigma}}[(X_j - \mu_j)^2] - 2E_{\vec{\mu}, \vec{\sigma}}[ \hat{B}(X_j - \mu_j)(X_j - \bar{X})] + E_{\vec{\mu}, \vec{\sigma}}[(\hat{B}(X_j - \bar{X}))^2]\\
    &= \calR_{\vec{\mu}, \vec{\sigma}}(\alg) - 2(m-3)[\Sigma_{j=1}^m E_{\vec{\mu}, \vec{\sigma}}[\frac{\hat{\sigma}^2(X_j - \mu_j)(X_j - \bar{X})}{S^2(\vec{X})}]] + (m-3)^2E_{\vec{\mu}, \vec{\sigma}}[\frac{\hat{\sigma}^4}{S^2(\vec{X})}]\\
    E_{\vec{\mu}, \vec{\sigma}}[\frac{\hat{\sigma}^2(X_j - \mu_j)(X_j - \bar{X})}{S^2(\vec{X})}] &= E_{\vec{\mu}, \vec{\sigma}}[g(X)(X_j - \mu_j)]\\
    &=_{stein} \sigma^2E_{\vec{\mu}, \vec{\sigma}}[\frac{d}{dX_j}g(X)]
    \end{align*}
    \begin{align*}
    \frac{d}{dX_j}g(X) &= \frac{d}{dX_j}\frac{\hat{\sigma}^2(X_j - \bar{X})}{S^2(\vec{X})}\\
    &= \frac{[(1-\frac{1}{m})\hat{\sigma}^2 + \frac{d\hat{\sigma}^2}{dX_j}(X_j - \bar{X})]S^2(\vec{X}) - 2\hat{\sigma}^2(X_j - \bar{X})^2}{S^4}\\
    &= \frac{\hat{\sigma}^2[(1-\frac{1}{m})S^2(\vec{X}) - 2(X_j - \bar{X})^2]+\frac{d\hat{\sigma}^2}{dX_j}(X_j - \bar{X})S^2(\vec{X})}{\big (S^2(\vec{X}) \big )^2}\\
    \Sigma_{j=1}^m E_{\vec{\mu}, \vec{\sigma}}[\frac{\hat{\sigma}^2(X_j - \mu_j)(X_j - \bar{X})}{S^2(\vec{X})}] &= \sigma^2(E_{\vec{\mu}, \vec{\sigma}}[\hat{\sigma}^2 \Sigma_{j=1}^m\frac{(1-\frac{1}{m})S^2(\vec{X}) - 2(X_j - \bar{X})^2}{\big (S^2(\vec{X}) \big )^2}] \\
    &~~~~~~~~+ E_{\vec{\mu}, \vec{\sigma}}[\frac{S^2(\vec{X}) \Sigma_{j=1}^m\frac{d\hat{\sigma}^2}{dX_j}(X_j - \bar{X})}{\big (S^2(\vec{X}) \big )^2}])\\
    &= \sigma^2[(m-3)E_{\vec{\mu}, \vec{\sigma}}[\frac{\hat{\sigma}^2}{S^2(\vec{X})}] + E_{\vec{\mu}, \vec{\sigma}}[\frac{\Sigma_{j=1}^m\frac{d\hat{\sigma}^2}{dX_j}(X_j - \bar{X})}{S^2(\vec{X})}]]
\end{align*}
Also mind that:
\begin{align*}
    E_{\vec{\mu}, \vec{\sigma}}[\frac{\hat{\sigma}^2(X-\mu)^T(X-\bar{X})}{S^2(\vec{X})}] &= cov(\hat{\sigma}^2,\frac{(X-\mu)^T(X-\bar{X})}{S^2(\vec{X})}) + E_{\vec{\mu}, \vec{\sigma}}[\hat{\sigma}^2]E_{\vec{\mu}, \vec{\sigma}}(\frac{(X-\mu)^T(X-\bar{X})}{S^2(\vec{X})}]\\
    &=_{stein} cov(\hat{\sigma}^2,\frac{(X-\mu)^T(X-\bar{X})}{S^2(\vec{X})}) + (m-3)\sigma^2E_{\vec{\mu}, \vec{\sigma}}[\hat{\sigma}^2]E_{\vec{\mu}, \vec{\sigma}}[\frac{1}{S^2(\vec{X})}]\\
    E_{\vec{\mu}, \vec{\sigma}}[\frac{\hat{\sigma}^2}{S^2(\vec{X})}] &= cov(\hat{\sigma}^2, \frac{1}{S^2(\vec{X})}) + E_{\vec{\mu}, \vec{\sigma}}[\hat{\sigma}^2]E_{\vec{\mu}, \vec{\sigma}}[\frac{1}{S^2(\vec{X})}]\\
\end{align*}
And therefore we get that:
\begin{align*}
    \sigma^2&E_{\vec{\mu}, \vec{\sigma}}[\frac{\Sigma_{j=1}^m\frac{d\hat{\sigma}^2}{dX_j}(X_j - \bar{X})}{S^2(\vec{X})}]] = cov(\hat{\sigma}^2,\frac{(X-\mu)^T(X-\bar{X})}{S^2(\vec{X})}) \\
    &~~~~+ (m-3)\sigma^2E_{\vec{\mu}, \vec{\sigma}}[\hat{\sigma}^2]E_{\vec{\mu}, \vec{\sigma}}[\frac{1}{S^2(\vec{X})}] - \sigma^2(m-3)E_{\vec{\mu}, \vec{\sigma}}[\frac{\hat{\sigma}^2}{S^2(\vec{X})}]\\
    &= cov(\hat{\sigma}^2,\frac{(X-\mu)^T(X-\bar{X})}{S^2(\vec{X})})\\ &~~~~+(m-3)\sigma^2E_{\vec{\mu}, \vec{\sigma}}[\hat{\sigma}^2]E_{\vec{\mu}, \vec{\sigma}}[\frac{1}{S^2(\vec{X})}] -\sigma^2(m-3)(cov(\hat{\sigma}^2, \frac{1}{S^2(\vec{X})}) + E_{\vec{\mu}, \vec{\sigma}}[\hat{\sigma}^2]E_{\vec{\mu}, \vec{\sigma}}[\frac{1}{S^2(\vec{X})}])\\
    &= cov(\hat{\sigma}^2,\frac{(X-\mu)^T(X-\bar{X})}{S^2(\vec{X})}) -\sigma^2(m-3)(cov(\hat{\sigma}^2, \frac{1}{S^2(\vec{X})}))\\
    &= cov(\hat{\sigma}^2,\frac{(X-\mu)^T(X-\bar{X}) - \sigma^2(m-3)}{S^2(\vec{X})})
\end{align*}

Plugging everything:
\begin{align*}
    \mathcal{R}_{\vec{\mu}, \vec{\sigma}}(\hat{\mu}, \mu) &= \calR_{\vec{\mu}, \vec{\sigma}}(\alg) - 2\sigma^2(m-3) \Big((m-3)E_{\vec{\mu}, \vec{\sigma}}[\frac{\hat{\sigma}^2}{S^2(\vec{X})}] + E_{\vec{\mu}, \vec{\sigma}}\Big[\frac{\Sigma_{j=1}^m\frac{d\hat{\sigma}^2}{dX_j}(X_j - \bar{X})}{S^2(\vec{X})}\Big]\Big)\\
    &~~~~+ (m-3)^2E_{\vec{\mu}, \vec{\sigma}}[\frac{\hat{\sigma}^4}{S^2(\vec{X})}]\\
    &= \calR_{\vec{\mu}, \vec{\sigma}}(\alg) + (m-3)^2\Big(E_{\vec{\mu}, \vec{\sigma}}[\frac{\hat{\sigma}^4}{S^2(\vec{X})}] - 2\sigma^2\Big[E_{\vec{\mu}, \vec{\sigma}}[\frac{\hat{\sigma}^2}{S^2(\vec{X})}] + E_{\vec{\mu}, \vec{\sigma}}[\frac{\Sigma_{j=1}^m\frac{d\hat{\sigma}^2}{dX_j}(X_j - \bar{X})}{(m-3)S^2(\vec{X})}]\Big]\Big)\\
    &= \calR_{\vec{\mu}, \vec{\sigma}}(\alg) + (m-3)^2(E_{\vec{\mu}, \vec{\sigma}}[\frac{\hat{\sigma}^4}{S^2(\vec{X})}] - 2\sigma^2E_{\vec{\mu}, \vec{\sigma}}[\frac{\hat{\sigma}^2}{S^2(\vec{X})}])\\
    &~~~~-2(m-3)cov\Big(\hat{\sigma}^2,\frac{(X-\mu)^T(X-\bar{X}) - \sigma^2(m-3)}{S^2(\vec{X})}\Big)
\end{align*}
\end{proof}
Denote by $\bar Y:=\frac{1}{m}\sum_jY_j$ and $S^2(Y):= \frac1{m-1}\sum_j(Y_j-\bar Y)^2$ its \emph{mean} and its \emph{sample variance}, respectively which yields the result.

\subsection{Corollary \ref{cor:meanAdjusted}}\label{proof:meanAdjusted}
Under the Normal model,  if $\estn$ is mean-adjusted then 
\begin{align*}
    &\mathcal{R}_{\vec{\mu}, \vec{\sigma}}(\ebAlg{\alg}{\estn}) < \mathcal{R}_{\vec{\mu}, \vec{\sigma}}(\alg) 
    \\ &+ 
    \frac{(m-3)^2}{m-1}(E_{\vec{\mu}, \vec{\sigma}}[\frac{(\estn(\vec X^A))^2}{S^2(\vec{X}^A)}] - 2{\sigma}^2E_{\vec{\mu}, \vec{\sigma}}[\frac{\estn(\vec X^A)}{S^2(\vec{X}^A)}])
\end{align*}

\begin{proof}
From theorem \ref{thm:single_worker_risk}:
$$\calR_{\vec{\mu}, \vec{\sigma}}(\ebAlg{\alg}{\estn}) = \calR_{\vec{\mu}, \vec{\sigma}}(\alg) +  \frac{(m-3)^2}{m-1}\Big(E_{\vec{\mu}, \vec{\sigma}}[\frac{(\estn(\vec X^A))^2}{S^2(\vec{X}^A)}]
  - 2{\sigma}^2\Big[E_{\vec{\mu}, \vec{\sigma}}[\frac{\estn(\vec X^A)}{S^2(\vec{X}^A)}] + E_{\vec{\mu}, \vec{\sigma}}\big[\frac{\Sigma_{j=1}^m\frac{d\psi(\vec X^A)}{dX^A_j}(X_j^A - \bar{X}^A)}{(m-3)S^2(\vec{X}^A)}\big]\Big]\Big)$$

Under the assumptions made it is easy to see that: 
\begin{align*}
    E_{\vec{\mu}, \vec{\sigma}}\big[\frac{\Sigma_{j=1}^m\frac{d\psi(\vec X^A)}{dX^A_j}(X_j^A - \bar{X}^A)}{(m-3)S^2(\vec{X}^A)}\big]\Big] &=  E_{\vec{\mu}, \vec{\sigma}}\big[\frac{\Sigma_{j=1}^m\frac{d\psi(\vec X^A)}{dX^A_j}(X_j^A - \bar{X}^A)}{(m-3)S^2(\vec{X}^A)}\big]1_{X_j^A \leq \bar{X}^A}\Big] \\
    &~~~+ E_{\vec{\mu}, \vec{\sigma}}\big[\frac{\Sigma_{j=1}^m\frac{d\psi(\vec X^A)}{dX^A_j}(X_j^A - \bar{X}^A)}{(m-3)S^2(\vec{X}^A)}\big]1_{X_j^A > \bar{X}^A}\Big] \\
    &> 0
\end{align*}
And the result immediately follows.
\end{proof}
\subsection{Additional Corollaries}\label{apx:additional_corollaries}
\begin{corollary}\label{indepcase}
If $\hat{\sigma}^2$ is independent of $X_j^A (\frac{d\hat{\sigma}^2}{dX_j^A} = 0) \ \forall j$ then directly from \ref{thm:single_worker_risk} we get that:
$$\calR_{\vec{\mu}, \vec{\sigma}}(\ebAlg{\alg}{\estn}) = \calR_{\vec{\mu}, \vec{\sigma}}(\alg) + E_{\vec{\mu},  \vec{\sigma}}[\frac{1}{S^2(\vec{X}^A)}](m-3)^2 (E_{\vec{\mu},  \vec{\sigma}}[\hat{\sigma}^4]-2\sigma^2E_{\vec{\mu},  \vec{\sigma}}[\hat{\sigma}^2])$$
and $\calR_{\vec{\mu}, \vec{\sigma}}(\ebAlg{\alg}{\estn}) < \calR_{\vec{\mu}, \vec{\sigma}}(\alg)$ if
\begin{align}
  \frac{E_{\vec{\mu},  \vec{\sigma}}[\hat{\sigma}^4]}{E_{\vec{\mu},  \vec{\sigma}}[\hat{\sigma}^2]} < 2\sigma^2  
\end{align}
\end{corollary}
\begin{corollary}
Assume $\hat{\sigma}^2$ is independent of $X_j \ \forall j$ and that $\exists \epsilon > 0$ $|\hat{\sigma}^2 - \sigma^2| < \epsilon$ w.p 1 then $\calR_{\vec{\mu}, \vec{\sigma}}(\ebAlg{\alg}{\estn}) < \calR_{\vec{\mu}, \vec{\sigma}}(\alg)$ if
\begin{equation}
\epsilon \in (0, \sigma^2) \  
\end{equation}
\end{corollary}
\begin{proof}
From corollary \ref{indepcase}, we found that $E_{\vec{\mu}, \vec{\sigma}}[\hat{\sigma}^4] < 2\sigma^2E_{\vec{\mu}, \vec{\sigma}}[\hat{\sigma}^2]$ and therefore:
\begin{align*}
    E_{\vec{\mu}, \vec{\sigma}}[\hat{\sigma}^4] &< (\sigma^2 + \epsilon)E_{\vec{\mu}, \vec{\sigma}}[\hat{\sigma}^2] <  2\sigma^2E_{\vec{\mu}, \vec{\sigma}}[\hat{\sigma}^2]\\
    (\sigma^2 + \epsilon) &< 2\sigma^2\\
    0 &< \epsilon < \sigma^2
\end{align*}
\end{proof}
\begin{corollary}\label{single_worker_corn}
Assume $\hat{\sigma}^2$ is independent of $X_j \ \forall j$ and that $\exists \epsilon > 0$ $|\hat{\sigma}^2 - \sigma^2| < \epsilon$ w.p $\delta$ and $\exists B$ s.t $\hat{\sigma}^2 < B$ w.p $1$ then $\calR_{\vec{\mu}, \vec{\sigma}}(\ebAlg{\alg}{\estn}) < \calR_{\vec{\mu}, \vec{\sigma}}(\alg) \ \forall \mu \in \mathbb{R}^m, m>3$ if
\begin{equation} \left\{
\begin{array}{@{}rl@{}}
B &< \frac{\delta}{1-\delta}\sigma^4\\
\epsilon &\in \big(0, -2\sigma^2 + \sqrt{5\sigma^4 + B(1 - \frac{1}{\delta})}\big)
\end{array}
\right .
\end{equation}
\end{corollary}
\begin{figure}
        \includegraphics[width=5cm]{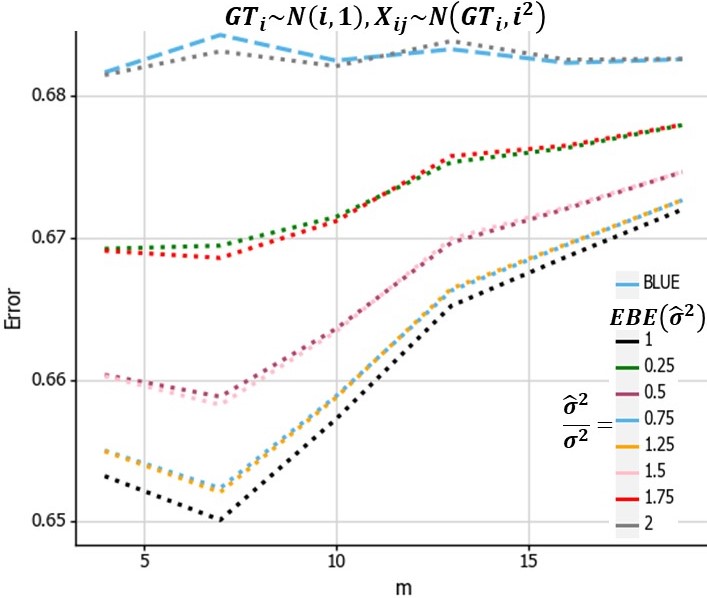}
        \centering
        \caption{EBE for 5 aggregated workers with biased $\hat{\sigma}^2$ vs BLUE; each data point is a 100,000 iteration average, each iteration includes new GT and new workers' responses}
\end{figure}
We know from corollary \ref{indepcase}, under the norml model, when the competence estimator is independent of the observations $X$, then EBE dominates BLUE if $E_{\vec{\mu}, \vec{\sigma}}[\hat{\sigma}^4] < 2\sigma^2E_{\vec{\mu}, \vec{\sigma}}[\hat{\sigma}^2]$.

Denote the event $G = \{|\sigma^2 - \hat{\sigma}^2| < \epsilon \}$
\begin{align*}
    E_{\vec{\mu}, \vec{\sigma}}[\hat{\sigma}^4] &= E_{\vec{\mu}, \vec{\sigma}}[\hat{\sigma}^4|G]\delta + E_{\vec{\mu}, \vec{\sigma}}[\hat{\sigma}^4|G^c](1-\delta)\\
    &\leq (\sigma^2 + \epsilon)^2\delta + B(1-\delta)\\
    2\sigma^2E_{\vec{\mu}, \vec{\sigma}}[\hat{\sigma}^2] &= 2\sigma^2E_{\vec{\mu}, \vec{\sigma}}[\hat{\sigma}^2|G]\delta + 2\sigma^2E_{\vec{\mu}, \vec{\sigma}}[\hat{\sigma}^2|G^c]\\
    &> 2\sigma^2(\sigma^2 - \epsilon)\delta
\end{align*}
Therefore it is sufficient to require that:
\begin{align*}
    (\sigma^2 + \epsilon)^2\delta + B(1-\delta) &< 2\sigma^2(\sigma^2 - \epsilon)\delta\\
    0 &< -\epsilon^2 -4\epsilon\sigma^2 + \sigma^4 + B(1 - \frac{1}{\delta})
\end{align*}
Which is a parabola of $\epsilon$ with roots $\epsilon=-2\sigma^2 \pm \sqrt{5\sigma^4 + B(1 - \frac{1}{\delta})}$. Notice that $B(1 - \frac{1}{\delta}) < 0$ and thus, we require that $2\sigma^2 < \sqrt{5\sigma^4 + B(1 - \frac{1}{\delta})}$ with simple algebra we derive the following conditions:
\begin{equation*}\left\{
\begin{array}{@{}rl@{}}
B &< \frac{\delta}{1-\delta}\sigma^4\\
\epsilon &\in \big(0, -2\sigma^2 + \sqrt{5\sigma^4 + B(1 - \frac{1}{\delta})}\big)\\
\end{array}
\right .
\end{equation*}
Notice when $\delta \xrightarrow{} 1$ we get that $\epsilon \in (0, \sigma^2(\sqrt{5} - 2))$, this upper bound is less than one forth of the case when $\delta=1$, this stricter result is due to the bounds we had to use to derive it.

\section{Deterministic Estimators for Multiple Workers}\label{apx:deterministic_est}
On this model we assume some oracle guessed and told us all of the different $\hat{\sigma}^2_i$ and thus, we treat them as  constants (i.e. independent of the data $\XX$), we would like to know how close the oracle has to be to the actual competences such that EBE would still have lower risk than estimated BLUE by some algorithm A, i.e $\calR_{\vec{\mu}, \vec{\sigma}}(\ebAlg{\alg}{\estn}) < \calR_{\vec{\mu}, \vec{\sigma}}(\alg)$.\par
$Let \ X_{ij} \sim \mathcal{N}(\mu_j, \sigma_i^2) \ i = 1,..,n \ j = 1,..,m $ and denote $\hat{\sigma}^2_i$ as an estimator which was somehow estimated for $\sigma^2_i, \forall i$.
Notice that the BLUE estimator (BLUE-aggregated worker) for $\mu_j$ is:
$$\hat{X}_j^A = (\Sigma_{i=1}^n\frac{1}{\hat{\sigma}^2_i})^{-1}\Sigma_{i=1}^n \frac{X_{ij}}{\hat{\sigma}^2_i}$$

 Then, notice that $\hat{X}_j^A$ is a linear combination of independent normal random variables therefore normal, i.e, $\hat{X}_j \sim N(\mu_j, \sigma^2)$ where:
\begin{align*}
    E_{\vec{\mu}, \vec{\sigma}}(\hat{X}_j^A) &= E_{\vec{\mu}, \vec{\sigma}}((\Sigma_{i=1}^n\frac{1}{\hat{\sigma}^2_i})^{-1}\Sigma_{i=1}^n \frac{X_{ij}}{\hat{\sigma}^2_i})= \mu_j\\
    Var_{\vec{\mu}, \vec{\sigma}}(\hat{X}_j^A) &= (\Sigma_{i=1}^n\frac{1}{\hat{\sigma}^2_i})^{-2}\Sigma_{i=1}^n \frac{\sigma^2}{\hat{\sigma}^4_i}
\end{align*}
Therefore, we can reduce this case to the case of single worker where: %\rmr{something messed up here:}
\begin{align*}
  \hat{\sigma}^2 = (\Sigma_{i=1}^n\frac{1}{\hat{\sigma}^2_i})^{-2}\Sigma_{i=1}^n \frac{\hat{\sigma}_i^2}{\hat{\sigma}^4_i} = (\Sigma_{i=1}^n\frac{1}{\hat{\sigma}^2_i})^{-1}\\  
  \calR_{\vec{\mu}, \vec{\sigma}}(\ebAlg{\alg}{\estn})_j =\hat{\bar{X}}^A + [1 - \frac{(m-3)\hat{\sigma}^2}{S^2(\vec{X}^A)}](\hat{X}_j^A - \hat{\bar{X}}^A)\\
  S^2(\vec{X}^A) = \sum_{j=1}^m(\hat{X}_j^A - \hat{\bar{X}}^A)^2
\end{align*}
\begin{proposition}
Under the Oracle model $\calR_{\vec{\mu}, \vec{\sigma}}(\ebAlg{\alg}{\estn}) < \calR_{\vec{\mu}, \vec{\sigma}}(\alg) \ \forall \mu \in \mathbb{R}^m, m>3$ if: 
\begin{equation}
\hat{\sigma}^2 < 2\sigma^2    
\end{equation}
\end{proposition}
\begin{proof}
Since we assumed an Oracle model (constant guesses of $\sigma^2$) and showed that $\hat{X}^A$ is following a normal distribution applying corollary \ref{indepcase} yields the result.
\end{proof}
% Recall the risk function from Corollary \ref{indepcase}: $$\calR_{\vec{\mu}, \vec{\sigma}}(\ebAlg{\alg}{\estn}) = \calR_{\vec{\mu}, \vec{\sigma}}(\alg) + E_{\vec{\mu},  \vec{\sigma}}[\frac{1}{S^2}](m-3)^2 (E_{\vec{\mu},  \vec{\sigma}}[\hat{\sigma}^4]-2\sigma^2E_{\vec{\mu},  \vec{\sigma}}[\hat{\sigma}^2])$$, under the oracle model, treating $\hat{\sigma}^2$ as a constant, we get the risk function:
% $$\mathcal{R}_{\vec{\mu}, \vec{\sigma}}(X,\hat{\mu}) + E_{\vec{\mu},  \vec{\sigma}}[\frac{1}{S^2}](m-3)^2 (\hat{\sigma}^4-2\sigma^2\hat{\sigma}^2)$$
% Observing the behaviour of the parabola  $h(\hat{\sigma}^2) =\hat{\sigma}^4-2\sigma^2\hat{\sigma}^2$ which we wish to minimize, we see that when $h(0) = h(2\sigma^2) = 0$ so we get that $\mathcal{R}_{\vec{\mu}, \vec{\sigma}}(\hat{\mu}, \mu)=\mathcal{R}_{\vec{\mu}, \vec{\sigma}}(X,\mu)$, i.e, BLUE and EBE coincide. Is it also easy to show that $h(\hat{\sigma}^2)$ has a minimum at $h(\sigma^2)=-\sigma^4$.

\end{small}

\end{document}